\documentclass{article}

\usepackage{arxiv}

\usepackage[american]{babel}

\usepackage{natbib} 
\usepackage{mathtools} 
\usepackage{booktabs} 
\usepackage{tikz} 

\usepackage[utf8]{inputenc}
\usepackage[T1]{fontenc} 
\usepackage{lmodern}
\usepackage{pifont}
\usepackage{amsmath,amsfonts,amssymb,dsfont,bm,mathtools,amsthm}
\usepackage{graphicx}
\usepackage{csquotes}
\usepackage[normalem]{ulem}
\usepackage{accents}
\usepackage{tikz}
\usetikzlibrary{shapes}
\usepackage{pgfplots}
\usetikzlibrary{graphs, graphs.standard}
\usepgfplotslibrary{statistics}
\usetikzlibrary{arrows}
\usetikzlibrary{patterns}
\usetikzlibrary{arrows,shapes,decorations,backgrounds}
\usepackage{algorithm}
\usepackage[noend]{algorithmic}
\usepackage{comment} 
\usepackage{adjustbox}

\usepackage{hyperref}
\usepackage{caption}
\usepackage{subcaption}
\usepackage{multirow}
\usepackage{colortbl}
\definecolor{Gray}{gray}{0.9}
\newcolumntype{a}{>{\columncolor{Gray}}c}
\usepackage{pifont}
\newcommand{\cmark}{\ding{51}}%
\newcommand{\xmark}{\ding{55}}%

\newtheorem{definition}{Definition}

\makeatother

\begin{document}

\title{Case Studies of Causal Discovery from IT Monitoring Time Series}

    \author{
 	\href{https://orcid.org/0009-0004-2715-7363}{\includegraphics[scale=0.06]{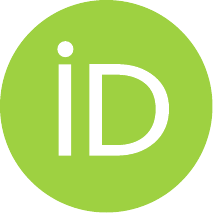}\hspace{1mm}Ali~Aït-Bachir} \\ EasyVista
   	\And
    \href{https://orcid.org/0000-0003-3571-3636}{\includegraphics[scale=0.06]{orcid.pdf}\hspace{1mm}Charles K. Assaad} \\ EasyVista 
	\And 
    \href{https://orcid.org/0009-0001-1687-797X}
    {\includegraphics[scale=0.06]{orcid.pdf}\hspace{1mm}Christophe~de~Bignicourt} \\ EasyVista
	\And
    \href{https://orcid.org/0000-0002-8360-1834}
    {\includegraphics[scale=0.06]{orcid.pdf}\hspace{1mm}Emilie~Devijver} \\Univ Grenoble Alpes, \\CNRS, Grenoble INP, LIG
	\And
    \href{https://orcid.org/0009-0001-4277-8377}{\includegraphics[scale=0.06]{orcid.pdf}\hspace{1mm}Simon~Ferreira} \\ École Normale Supérieure de Lyon, \\EasyVista
    \And
    \href{https://orcid.org/0000-0002-8858-3233}{\includegraphics[scale=0.06]{orcid.pdf}\hspace{1mm}Eric Gaussier} \\Univ Grenoble Alpes, \\CNRS, Grenoble INP, LIG
    \And
    \href{https://orcid.org/0009-0008-7791-0092}{\includegraphics[scale=0.06]{orcid.pdf}\hspace{1mm}Hosein~Mohanna} \\ EasyVista
    \And
    \href{https://orcid.org/0000-0003-4695-5059}{\includegraphics[scale=0.06]{orcid.pdf}\hspace{1mm}Lei~Zan} \\ Univ Grenoble Alpes, \\CNRS, Grenoble INP, LIG, \\ EasyVista
}
\def\thefootnote{}\footnotetext{Authors are listed in an alphabetical oder.}\def\thefootnote{\arabic{footnote}}

\date{}

\maketitle

\begin{abstract}
	Information technology (IT) systems are vital for modern businesses, handling data storage, communication, and process automation. Monitoring these systems is crucial for their proper functioning and efficiency, as it allows collecting extensive observational time series data for analysis. 
	The interest in causal discovery is growing in IT monitoring systems as knowing causal relations between different components of the IT system  helps in reducing downtime, enhancing system performance and identifying root causes of anomalies and incidents. It also allows proactive prediction of future issues through historical data analysis. Despite its potential benefits, applying causal discovery algorithms on IT monitoring data poses challenges, due to the complexity of the data. For instance, IT monitoring data often contains misaligned time series, sleeping time series, timestamp errors and missing values. This paper presents case studies on applying causal discovery algorithms to different IT monitoring datasets, highlighting benefits and ongoing challenges.
	
\end{abstract}

\section{Introduction}

Information technology (IT) systems play a crucial role in the success of modern businesses. These systems are utilized for data storage and processing, communication with customers and suppliers, and the automation of various business processes. Given their significance, it is essential to monitor IT systems to ensure their proper functioning and efficiency \citep{Tamburri_2020}. IT monitoring has become increasingly valuable due to improved storage capacity, enabling the collection of extensive observational time series data \citep{Tamburri_2020}.
Even though analyzing these large amounts of observational time series data can enhance efficiency and optimize processes \citep{Chatzigiannakis_2009}, they also pose a significant challenge for many companies due to their complex nature.

The interest in causal discovery~\citep{Spirtes_2000,Pearl_2000,Chickering_2002,Peters_2017}  is growing within the IT monitoring community~\citep{Meng_2020,Li_2022,Assaad_2023,Wang_2023}, as knowing causal relations allows for reducing downtime and enhancing the overall performance of IT systems by optimizing their  resources and identifying areas for improvement. In addition, causal discovery can help IT professionals to swiftly identify actionable root causes of anomalies and incidents and to take corrective action to eliminate them~\citep{Meng_2020,Assaad_2023}.
Moreover, causal discovery can also be used to predict and preempt future issues in IT systems. By analyzing historical data, IT professionals can recognize patterns indicative of future problems and address them proactively. By leveraging causal discovery, IT professionals can enhance the efficiency, performance, and reliability of IT systems, leading to improved business outcomes.

However, analyzing IT monitoring data poses several challenges due to its complexity. IT monitoring data is often collected from multiple sources, resulting in misaligned time series. Additionally, there can also be non-informative time series. For example, there can be sleeping time series (for a period of time) due to users' inactivity on certain servers.
Furthermore, timestamp errors can be present, and the low sampling rate of data in IT monitoring systems can complicate the search for causal relationships, as the lag between causes and effects may be relatively small.

This paper presents a case study for applying causal discovery algorithms to different IT monitoring datasets. This study highlights the potential benefits of utilizing causal discovery techniques in IT monitoring and  emphasizes the ongoing challenges and complexities associated with working with such data. These findings stress on the need for further research and development in this area to fully harness the potential of causal discovery algorithms in analyzing IT monitoring data.

The remainder of the paper is organized as follows: Section~\ref{sec:setup} presents preliminaries and the main algorithms applied to the case studies.
Section~\ref{sec:datasets} describes the IT monitoring dataset for each case study and discusses the challenges related to each dataset as well as the  background knowledge and theories available to experts at the time of the application.
Section~\ref{sec:results} presents and discusses the results of causal discovery algorithm in each case study.
Finally, Section~\ref{sec:elaboration} discusses challenges and points
out some aspect of causal discovery from time series that are not included in any of the case studies and Section~\ref{sec:conclusion} concludes the paper.

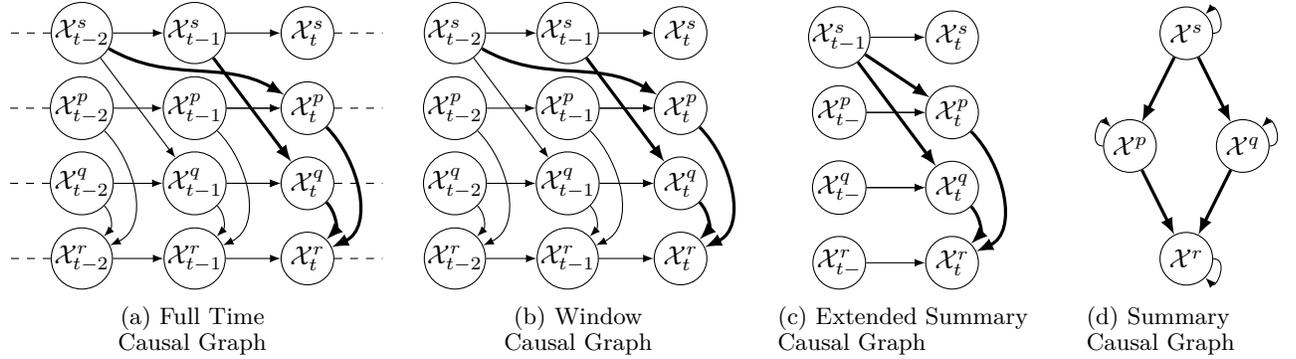
\begin{figure*}[t!]
	\centering
	\begin{subfigure}{.3\textwidth}
		\begin{tikzpicture}[{black, circle, draw, inner sep=0}]
		\tikzset{nodes={draw,rounded corners},minimum height=0.7cm,minimum width=0.7cm}
		\tikzset{latent/.append style={fill=gray!30}}
		
		\node (X-2) at (0,2) {$\mathcal{X}^s_{t-2}$} ;
		\node (X-1) at (1.5,2) {$\mathcal{X}^s_{t-1}$};
		\node (X) at (3,2) {$\mathcal{X}^s_{t}$};
		\node (Z-2) at (0,1) {$\mathcal{X}^p_{t-2}$} ;
		\node (Z-1) at (1.5,1) {$\mathcal{X}^p_{t-1}$};
		\node (Z) at (3,1) {$\mathcal{X}^p_{t}$};
		\node (Y-2) at (0,0) {$\mathcal{X}^q_{t-2}$} ;
		\node (Y-1) at (1.5,0) {$\mathcal{X}^q_{t-1}$};
		\node (Y) at (3,0) {$\mathcal{X}^q_{t}$};
		\node (W-2) at (0,-1) {$\mathcal{X}^r_{t-2}$} ;
		\node (W-1) at (1.5,-1) {$\mathcal{X}^r_{t-1}$};
		\node (W) at (3,-1) {$\mathcal{X}^r_{t}$};
		
		\draw[->,>=latex] (Z-2) -- (Z-1);
		\draw[->,>=latex] (Z-1) -- (Z);
		\draw[->,>=latex] (Y-2) -- (Y-1);
		\draw[->,>=latex] (Y-1) -- (Y);
		\draw[->,>=latex] (X-2) -- (X-1);
		\draw[->,>=latex] (X-1) -- (X);
		\draw[->,>=latex] (W-2) -- (W-1);
		\draw[->,>=latex] (W-1) -- (W);
		
		\draw[->,>=latex, very thick] (X-2) to [out=-30,in=-210, looseness=1] (Z);
		\draw[->,>=latex] (X-2) -- (Y-1);
		\draw[->,>=latex, very thick] (X-1) -- (Y);

		\draw[->,>=latex] (Z-1) -- (Z);
		\draw[->,>=latex] (Y-1) -- (Y);
		
		\draw[->,>=latex, very thick] (Y) to [out=-45,in=45, looseness=1] (W);
		\draw[->,>=latex, very thick] (Z) to [out=-45,in=25, looseness=1] (W);		
		\draw[->,>=latex] (Y-1) to [out=-45,in=45, looseness=1] (W-1);
		\draw[->,>=latex] (Z-1) to [out=-45,in=25, looseness=1] (W-1);	
		\draw[->,>=latex] (Y-2) to [out=-45,in=45, looseness=1] (W-2);
		\draw[->,>=latex] (Z-2) to [out=-45,in=25, looseness=1] (W-2);
		
		\coordinate[left of=X-2] (d1);
		\draw [dashed,>=latex] (X-2) to[left] (d1);
		\coordinate[left of=Z-2] (d1);
		\draw [dashed,>=latex] (Z-2) to[left] (d1);
		\coordinate[left of=Y-2] (d1);
		\draw [dashed,>=latex] (Y-2) to[left] (d1);		
		\coordinate[left of=W-2] (d1);
		\draw [dashed,>=latex] (W-2) to[left] (d1);
		
		\coordinate[right of=X] (d1);
		\draw [dashed,>=latex] (X) to[right] (d1);
		\coordinate[right of=Z] (d1);
		\draw [dashed,>=latex] (Z) to[right] (d1);
		\coordinate[right of=Y] (d1);
		\draw [dashed,>=latex] (Y) to[right] (d1);
		\coordinate[right of=W] (d1);
		\draw [dashed,>=latex] (W) to[right] (d1);
		\end{tikzpicture}
		\caption{Full Time \\Causal Graph}
		\label{fig:full-graph}
	\end{subfigure}
	\hfill
	\begin{subfigure}{.25\textwidth}
		\begin{tikzpicture}[{black, circle, draw, inner sep=0}]
		\tikzset{nodes={draw,rounded corners},minimum height=0.7cm,minimum width=0.7cm}
		\tikzset{latent/.append style={fill=gray!30}}
		
		\node (X-2) at (0,2) {$\mathcal{X}^s_{t-2}$} ;
		\node (X-1) at (1.5,2) {$\mathcal{X}^s_{t-1}$};
		\node (X) at (3,2) {$\mathcal{X}^s_{t}$};
		\node (Z-2) at (0,1) {$\mathcal{X}^p_{t-2}$} ;
		\node (Z-1) at (1.5,1) {$\mathcal{X}^p_{t-1}$};
		\node (Z) at (3,1) {$\mathcal{X}^p_{t}$};
		\node (Y-2) at (0,0) {$\mathcal{X}^q_{t-2}$} ;
		\node (Y-1) at (1.5,0) {$\mathcal{X}^q_{t-1}$};
		\node (Y) at (3,0) {$\mathcal{X}^q_{t}$};
		\node (W-2) at (0,-1) {$\mathcal{X}^r_{t-2}$} ;
		\node (W-1) at (1.5,-1) {$\mathcal{X}^r_{t-1}$};
		\node (W) at (3,-1) {$\mathcal{X}^r_{t}$};
		
		\draw[->,>=latex] (Z-2) -- (Z-1);
		\draw[->,>=latex] (Z-1) -- (Z);
		\draw[->,>=latex] (Y-2) -- (Y-1);
		\draw[->,>=latex] (Y-1) -- (Y);
		\draw[->,>=latex] (X-2) -- (X-1);
		\draw[->,>=latex] (X-1) -- (X);
		\draw[->,>=latex] (W-2) -- (W-1);
		\draw[->,>=latex] (W-1) -- (W);
		
		\draw[->,>=latex, very thick] (X-2) to [out=-30,in=-210, looseness=1] (Z);
		\draw[->,>=latex] (X-2) -- (Y-1);
		\draw[->,>=latex, very thick] (X-1) -- (Y);

		\draw[->,>=latex] (Z-1) -- (Z);
		\draw[->,>=latex] (Y-1) -- (Y);
		
		\draw[->,>=latex, very thick] (Y) to [out=-45,in=45, looseness=1] (W);
		\draw[->,>=latex, very thick] (Z) to [out=-45,in=25, looseness=1] (W);		
		\draw[->,>=latex] (Y-1) to [out=-45,in=45, looseness=1] (W-1);
		\draw[->,>=latex] (Z-1) to [out=-45,in=25, looseness=1] (W-1);	
		\draw[->,>=latex] (Y-2) to [out=-45,in=45, looseness=1] (W-2);
		\draw[->,>=latex] (Z-2) to [out=-45,in=25, looseness=1] (W-2);
		\end{tikzpicture}
		\caption{Window \\Causal Graph}
		\label{fig:window-graph}
	\end{subfigure}
	\hfill 
	\begin{subfigure}{.22\textwidth}
		\centering
		\begin{tikzpicture}[{black, circle, draw, inner sep=0}]
		\tikzset{nodes={draw,rounded corners},minimum height=0.7cm,minimum width=0.7cm}
		\tikzset{latent/.append style={fill=gray!30}}
		
		\node (X-1) at (0,2) {$\mathcal{X}^s_{t-1}$};
		\node (X) at (1.5,2) {$\mathcal{X}^s_{t}$};
		\node (Z-1) at (0,1) {$\mathcal{X}^p_{t-}$};
		\node (Z) at (1.5,1) {$\mathcal{X}^p_{t}$};
		\node (Y-1) at (0,0) {$\mathcal{X}^q_{t-}$};
		\node (Y) at (1.5,0) {$\mathcal{X}^q_{t}$};
		\node (W-1) at (0,-1) {$\mathcal{X}^r_{t-}$};
		\node (W) at (1.5,-1) {$\mathcal{X}^r_{t}$};
		
		\draw[->,>=latex, very thick] (Y) to [out=-45,in=45, looseness=1] (W);
		\draw[->,>=latex, very thick] (Z) to [out=-45,in=25, looseness=1] (W);		
		
		\draw[->,>=latex] (Z-1) -- (Z);
		\draw[->,>=latex] (Y-1) -- (Y);
		\draw[->,>=latex] (X-1) -- (X);
		\draw[->,>=latex] (W-1) -- (W);
		\draw[->,>=latex, very thick] (X-1) -- (Z);
		\draw[->,>=latex, very thick] (X-1) -- (Y);
		
		\draw[->,>=latex] (Z-1) -- (Z);
		\draw[->,>=latex] (Y-1) -- (Y);
		\end{tikzpicture}
		\caption{Extended Summary \\ Causal Graph}
		\label{fig:extended-graph}
	\end{subfigure}\hfill
	\begin{subfigure}{.13\textwidth}
		\centering
		\begin{tikzpicture}[{black, circle, draw, inner sep=0}]
		\tikzset{nodes={draw,rounded corners},minimum height=0.7cm,minimum width=0.7cm}
		\tikzset{latent/.append style={fill=gray!30}}
		
		\node (X) at (0,0.5) {$\mathcal{X}^p$} ;
		\node (Z) at (0.75,2) {$\mathcal{X}^s$};
		\node (Y) at (1.5,0.5) {$\mathcal{X}^q$};
		\node (W) at (0.75,-1) {$\mathcal{X}^r$};
		\draw[->,>=latex] (Y) to [out=0,in=45, looseness=2] (Y);
		\draw[->,>=latex] (Z) to [out=0,in=45, looseness=2] (Z);
		\draw[->,>=latex] (X) to [out=180,in=135, looseness=2] (X);
		\draw[->,>=latex] (W) to [out=0,in=-45, looseness=2] (W);
		
		\draw[->,>=latex, very thick] ( Z) -- (X);
		\draw[->,>=latex, very thick] (Z) -- (Y);
		
		\draw[<-,>=latex, very thick] (W) -- (X);
		\draw[<-,>=latex, very thick] (W) -- (Y);
		\end{tikzpicture}
		\caption{Summary \\Causal Graph}
		\label{fig:summary-graph}
	\end{subfigure}
	\caption{Different causal graphs to respresent a diamond structure with self causes: full time causal graph (a), window causal graph (b), extended summary causal graph (c) and summary causal graph (d). Note that the first one gives more information but cannot be inferred in practice, the second one is a schematic viewpoint of the full behavior, the third one only distinguishes between instantaneous and lagged causal relations, whereas the last one gives an overview of the causal relationships without any reference to time.}
	\label{fig:diamond}
\end{figure*}%

\section{Set up}
\label{sec:setup}
Causal discovery in time series aims at discovering, from observational data, causal relations within and between $d$-variate time series $\mathcal{X}$ where, for a fixed $t$, each $\mathcal{X}_t$ is a vector $(\mathcal X_t^1,\cdots,\mathcal X_t^d)$ in which each variable $\mathcal X_t^p$, such that $p\in \{1, \cdots, d\}$, represents a measurement of the $p$-th time series at time $t$. 

\subsection{Causal graphs for time series}
\label{sec:setup:graphs}
There are at least four ways to represent time series through a causal graph. 
The first is called a \emph{full time causal graph} \citep{Assaad_2022} and represents a infinite graph of the dynamic system, as illustrated in Figure \ref{fig:full-graph}. Note that in this work, we \emph{assume} that the \emph{full time causal graph} is \emph{acyclic}.

\begin{definition}[Full time causal graph, \cite{Assaad_2022}]
	Let $\mathcal{X}$ be a multivariate discrete-time stochastic process and $\mathcal{G}^f = (\mathcal{V}^f, \mathcal{E}^f)$ the associated \emph{full time causal graph}. The set of vertices $\mathcal{V}^f$ in that graph consists of the set of components $\mathcal{X}^1_t,\ldots, \mathcal{X}^d_t$ at each time $t\in \mathbb{Z}$. The set of edges $\mathcal{E}^f$ of the graph are defined as follows: for each $t$, variables $\mathcal X^p_{t-\gamma}$ and $\mathcal X^q_t$ are connected by a lag-specific directed link $\mathcal X^p_{t-\gamma} \rightarrow \mathcal X^q_t$ if and only if $\mathcal X^p$ causes $\mathcal X^q$ at time $t$ with a time lag of $0 \leq \gamma$ for $p\ne q$ and with a time lag of $0 < \gamma$ for $p= q$.
\end{definition}
\noindent It is usually not possible to infer general full time causal graphs as there usually is a single observation for each time series at each time instant. Thus it is common to rely on the so-called \emph{consistency throughout time assumption} \citep{Assaad_2022} which states that all causal relationships remain constant in direction throughout time. When assuming consistency throughout time and because every causal relation has a maximal time lag $\gamma_{max}$, the full time causal graph can be represented through a time window by a finite graph of size $\gamma_{max}+1$ which we call \emph{window causal graph} \citep{Assaad_2022}.

\begin{definition}[Window  causal graph, \cite{Assaad_2022}]
	Let $\mathcal{X}$ be a multivariate discrete-time stochastic process and $\mathcal{G}^w = (\mathcal{V}^w, \mathcal{E}^w)$ the associated \emph{window  causal graph} with a maximal lag $\gamma_{max}$. The set of vertices $\mathcal{V}^w$ in that graph consists of the set of components $\mathcal{X}^1_{t-\gamma},\ldots, \mathcal{X}^d_{t-\gamma}$ at each time $t-\gamma$ for $0\leq \gamma \leq \gamma_{max}$. The set of edges $\mathcal{E}^w$ of the graph are defined as follows: $\mathcal X^p_{t-\gamma}$ and $\mathcal X^q_{t}$ are connected by a directed link $\mathcal X^p_{t-\gamma} \rightarrow \mathcal X^q_{t}$ if and only if $\mathcal X^p_{t-\gamma}$ causes $\mathcal X^q_t$ in the full time causal graph (in this case then there is also a directed edge between each homologous pairs of nodes $\mathcal X^p_{t-\gamma-i}$ and $\mathcal X^q_{t-i}$ for $0 \le i \le \gamma_{max}-\gamma$). 
\end{definition}

\noindent Figure \ref{fig:window-graph} illustrates a window causal graph corresponding to the full time causal graph given in Figure \ref{fig:full-graph}.

In practice, it can be sufficient to know the causal relations between time series as a whole, without knowing precisely the relations between time instants; in addition, in some applications, an expert would like to validate a causal graph before using it, but validating a window causal graph and its temporal lags between causes and effects can be difficult. 
In these cases, one can use an abstraction of the window graph which usually takes the form of an \emph{extended summary causal graph} \citep{Assaad_2022b} or a \emph{summary causal graph} \citep{Assaad_2022}. An example of these two abstract graphs are given in Figure~\ref{fig:extended-graph} and \ref{fig:summary-graph}.

\begin{definition}[Extended summary causal graph, \cite{Assaad_2022b}]
	\label{Ext_Summary_G}
	Let $\mathcal{X}$ be a multivariate discrete-time stochastic process and $\mathcal{G}^e = (\mathcal{V}^e, \mathcal{E}^e)$ the associated \emph{extended summary causal graph}. The set of vertices $\mathcal{V}^e$ in that graph consists of the set of time slices $\mathcal V^e_{t-}$ and $\mathcal V^e_t$ such that $\mathcal V^e_{t-}= \mathcal{X}^1_{t-},\ldots, \mathcal{X}^d_{t-}$ and $V^e_{t}=\mathcal{X}^1_t,\ldots, \mathcal{X}^d_t$.
	The set of edges $\mathcal{E}^e$ are defined as follows:
	\begin{itemize}
		\item  variables $\mathcal X^p_{t}$ and $\mathcal X^q_{t}$ with $p \ne q$ are connected by a directed link $\mathcal X^p_{t} \rightarrow \mathcal X^q_{t}$ if and only if $\mathcal X^p$ causes $\mathcal X^q$ at time $t$ with a null time lag;
		\item variables $\mathcal X^p_{t-}$ and $\mathcal X^q_{t}$ are connected by a directed link $\mathcal X^p_{t-} \rightarrow \mathcal X^q_{t}$, if and only if $\mathcal X^p$ causes $\mathcal X^q$ at time $t$ with a strictly positive time lag.
	\end{itemize}
\end{definition}

\begin{definition}[Summary causal graph, \cite{Assaad_2022}]
	Let $\mathcal{X}$ be a multivariate discrete-time stochastic process and $\mathcal{G} = (\mathcal{V}, \mathcal{E})$ the associated \emph{summary causal graph}. The set of vertices $\mathcal{V}$ in that graph consists of the set of time series $\mathcal{X}^1,\ldots, \mathcal{X}^d$. The set of edges $\mathcal{E}$ of the graph are defined as follows: variables $\mathcal X^p$ and $\mathcal X^q$ are connected if and only if there exists some time $t$ and some time lag $i$ such that $\mathcal X^p_{t-i}$ causes $\mathcal X^q_{t}$ at time $t$ with a time lag of $0\leq i$ for $p \ne q$ and with a time lag of $0<i$ for $p=q$.
\end{definition}

Note that the summary causal graph can contain cycles which is not the case for extended summary causal graphs.

\begin{table*}[t!]
	\centering
	\begin{tabular}{cc@{\hspace{0.3cm}}acccccccaaaaaa}
		&&
		\rotatebox{90}{Causal graph} &  \rotatebox{90}{Causal Markov Condition} & \rotatebox{90}{Causal sufficiency} &
		\rotatebox{90}{Faithfulness / Minimality}  &
		\rotatebox{90}{Semi-parametric model}& \rotatebox{90}{Linear  model}. & \rotatebox{90}{Consistency throughout time} &  \rotatebox{90}{Stationarity}
		&\rotatebox{90}{Instantaneous relations}
		& \rotatebox{90}{Misaligned time series}  &
		\rotatebox{90}{Sleeping time series} & \rotatebox{90}{Timestamp errors} & \rotatebox{90}{Missing values} & \rotatebox{90}{Different sampling rate}\\
		\hline
		\multirow{8}{*}{\rotatebox{90}{Algorithms}}&GCMVL & S/E & \cmark & \cmark &  & \cmark & \cmark & \cmark  &\cmark & \xmark & \xmark  & \xmark & \xmark & \xmark & \xmark \\ 
		&Dynotears & W &  \cmark & \cmark &  & \cmark & \cmark & \cmark  &\cmark & \cmark & \xmark  & \xmark & \xmark & \xmark & \xmark \\ 	
		&PCMCI$^+$ & W&  \cmark & \cmark & F & \xmark & \xmark & \cmark  &\cmark & \cmark & \xmark  & \xmark & \xmark & \xmark & \xmark \\
		&PCGCE &E &  \cmark & \cmark &  F & \xmark & \xmark & \cmark  &\cmark & \cmark & \xmark  & \xmark & \xmark & \xmark & \xmark \\
		&VarLiNGAM & W &  \cmark & \cmark & M  & \cmark & \cmark & \cmark  &\cmark & \cmark & \xmark  & \xmark & \xmark & \xmark & \xmark \\ 	
		&TiMINo & S &  \cmark & \cmark & M & \cmark & \xmark & \cmark  &\cmark & \cmark & \xmark  & \xmark & \xmark & \xmark & \xmark \\ 	
		&NBCB-w & S/W & \cmark & \cmark & M & \cmark & \xmark & \cmark  &\cmark & \cmark & \xmark  & \xmark & \xmark & \xmark & \xmark \\ 
		&NBCB-e & S/W & \cmark & \cmark & M & \cmark & \xmark & \cmark  &\cmark & \cmark & \xmark  & \xmark & \xmark & \xmark & \xmark \\ 
		&CBNB-w & S/W & \cmark & \cmark & M & \cmark & \xmark & \cmark  &\cmark & \cmark & \xmark  & \xmark & \xmark & \xmark & \xmark \\ 
		&CBNB-e & S/W & \cmark & \cmark & M & \cmark & \xmark & \cmark  &\cmark & \cmark & \xmark  & \xmark & \xmark & \xmark & \xmark \\ 
		
		\hline
		\hline
		\multirow{4}{*}{\rotatebox{90}{Datasets}}&	MoM & S & \cmark & \cmark & ?& ? &? & \xmark & \xmark & \cmark & \xmark & \cmark & \cmark & \xmark & \xmark \\
		&Ingestion & S & \cmark & \cmark &? & ? &?  & \xmark & \xmark  & \cmark & \xmark & \cmark & \cmark & \xmark & \xmark \\ 
		&Web activity & S & \cmark & ? &? & ? &?  & \xmark & \xmark  & \cmark & \cmark & \cmark & \cmark & \xmark & \xmark \\ 
		& Antivirus & S& \cmark& ?& ? &  ? &? & \xmark & \xmark & \cmark & \cmark & \cmark & \cmark & \xmark & \cmark 
	\end{tabular}
	\caption{\label{tab:summary_algo} Summary of the main characteristics of  algorithms  and different IT monitoring datasets considered in the paper.  For causal graphs, S means that the algorithm provides a summary causal graph, E means that the algorithm provides an extended summary causal graph and W means that the algorithm provides a window causal graph; F corresponds to faithfulness and M to minimality.
		An empty cell mean that the information given in the corresponding column was not discussed by the authors of the corresponding algorithm. A question mark means that the expert of the IT system do not know if the information given in the corresponding column is satisfied for the given dataset.}
\end{table*}

\subsection{Assumptions}

Given observational data,  on which one can compute correlations and statistical independencies, it is not always possible to infer a causal graph. 
In addition to the acyclicity of the full time causal graph and consistency throughout time, all the algorithms considered in this work rely on some of the following assumptions:
\begin{itemize}
	\item Causal Markov condition \citep{Spirtes_2000,Pearl_2000}: every variable is independent of all its nondescendants in the graph conditional on its parents;
	\item Causal sufficiency \citep{Spirtes_2000,Pearl_2000}: all common causes, i.e., confounders, of all observed variables are observed;
	\item Minimality \citep{Spirtes_2000}: all adjacent nodes are dependent;
	\item Faithfulness \citep{Spirtes_2000,Pearl_2000}: all conditional independencies are entailed from the causal Markov condition;
	\item Semi-parametric model \citep{Peters_2017}, which stipulates a general form for the underlying model, as linear models or nonlinear additive noise models;  
	\item Stationarity: the generative process does not change with respect to time.
\end{itemize}

\subsection{Algorithms}

Granger Causality is one of the oldest methods proposed to detect causal relations between time series. However, in its standard form \citep{Granger_1969}, it is known to handle a restricted version of causality that focuses on linear relations and causal priorities as it assumes that the past of a cause is necessary and sufficient for optimally forecasting its effect. This approach has nevertheless been improved since then \citep{Granger_2004, Arnold_2007} through, \textit{e.g}, the use of variable selection tools and result of this method can be represented in the form of a summary causal graph. Namely,  GCMVL~\citep{Arnold_2007} is multivariate Granger algorithm that use a lasso-based technique for variable selection.


Score-based approaches \citep{Chickering_2002} search over the space of possible graphs trying to maximize a score that reflects how well the graph fits the data.
Recently, a new score-based method called Dynotears\footnote{\url{https://github.com/quantumblacklabs/causalnex/}} \citep{Pamfil_2020} was presented to infer a window causal graph from time series.

Constraint-based approaches, based on the PC algorithm  \citep{Spirtes_2000}, are certainly one of the most popular approaches for inferring causal graphs. Several algorithms, adapted from non-temporal causal graph discovery algorithms, have been proposed in this family for time series, among which PCMCI is capable of infering a window causal graph and accounts for the effect size.  Initially, PCMCI~\citep{Runge_2019} was not able to take into account instantaneous relations but this limitation was recently surmounted with the introduction of PCMCI$^+$\footnote{\url{https://github.com/jakobrunge/tigramite}}\citep{Runge_2020}. Another algorithm in this family is PCGCE\footnote{\url{https://github.com/ckassaad/PCGCE}}~\citep{Assaad_2022b} which infers an extended summary causal graph by restructuring the data into two slices: one vector for each time series that represents the present and one matrix of each time series that represents the past (up to $\gamma_{max}$).

In a different line, approaches based on Structural Equation Models assume that the causal system can be defined by a set of equations that explain each variable by its direct causes and an additional noise. Causal relations are in this case discovered using footprints produced by the causal asymmetry in the data. For time series, the most  popular algorithms in this family are VarLiNGAM\footnote{\url{https://github.com/cdt15/lingam}} ~\citep{Hyvarinen_2008,Hyvarinen_2010}, which is an extension of LiNGAM~\citep{Shimizu_2011} through autoregressive models that infers a window causal graph, and TiMINo\footnote{\url{http://web.math.ku.dk/~peters/code.html}}~\citep{Peters_2013}, which discovers a causal relationship in form of a summary causal graph by looking at independence between the noise and the potential causes.

There exist also hybrid algorithms which combine constraint-based with semi-parametric algorithms.  Among hybrid methods, NBCB~\citep{Assaad_2021} starts by discovering the causal order between time series through a semi-parametric strategy (which yields a graph that contains the true graph), and then prunes unnecessary edges using a constraint-based strategy.
Initially, this method assumes that the summary causal graph is acyclic but this limitation was recently surmounted~\citep{Assaad_2023_b}. In addition, in the new generalized version, NBCB is considered more like a framework that combines any semi-parametric strategy and constraint-based strategy. In this paper, we consider NBCB-w\footnote{\url{https://github.com/ckassaad/Hybrids_of_CB_and_NB_for_Time_Series}} which combines a restricted version of VarLiNGAM and a restricted version of PCMCI$^+$ as well as NBCB-e\footnotemark[\value{footnote}] which combines a restricted version of VarLiNGAM and a restricted version of PCGCE. NBCB-w infers a window causal graph as it is based on PCMCI$^+$ and NBCB-e infer an extended summary causal graph as it is based on PCGCE.
Another hybrid-based framework exists and it is called CBNB~\citep{Assaad_2023_b}. It can be considered as a backward version of NBCB. 
In this paper, we consider the two algorithms CBNB-w\footnotemark[\value{footnote}] and CBNB-e\footnotemark[\value{footnote}] from the CBNB framework  which can be considered respectively as the backward versions of NBCB-w and NBCB-e.

In Table~\ref{tab:summary_algo}, we classify causal discovery algorithms with respect to the assumptions they rely on in addition to different characteristics. 

\subsection{Hyper-parametres}
For PCMCI$^+$, PCGCE, NBCB-w, NBCB-e, CBNB-w, and CBNB-e we use the linear partial correlation to find conditional independencies and for PCGCE (as well as for NBCB-e and CBNB-e), as the authors suggested~\citep{Assaad_2022b}, we reduce the dimensionality of the past slice to $1$ using PCA.
For TiMINo~\citep{Peters_2013} we use the linear time series model and the HSIC test and for VLiNGAM, the regularization parameter in the adaptive Lasso is selected using BIC.
For Dynotears, we set all hyperparameters to their recommended values ($\lambda_W = \lambda_A = 0.05$ and $\alpha_W = \alpha_A = 0.01$).
For all methods, we set the significance threshold to $0.05$ since according to IT monitoring experts the maximal delay between a cause and its effect is of $15$ minutes,  in our experiments we set the maximal lag $\gamma_{max}$ according to the sampling rate and to the $15$ minutes delay. For instance, for a sampling rate of $1$ minute we set $\gamma_{max}$ to $15$ and for a sampling rate of $5$ minute we set $\gamma_{max}$ to $3$.  
In Appendix, we also study how the results change by varying $\gamma_{max}$.

\subsection{Pre-processing}
\label{sec:pre-process}
Time series in monitoring systems are not always exactly aligned together and come in different sampling rates as the timestamps depend on when the data was collected. In the following we present two pre-processing strategies that we considered for aligning time series:

\begin{itemize}
	\item Strategy~1:  Time series are analyzed in terms of sampling rates and the lowest one is chosen. Afterwards, all the time series are re-sampled according to this lowest sampling rate with the closest value to the timestamp taken as the new value. Upon re-sampling, missing values can be clearly observed. If missing values are detected, they are filled using simple linear interpolation of Pandas data frames\footnote{\url{https://pandas.pydata.org/docs/reference/api/pandas.DataFrame.interpolate.html}}.
	
	\item Strategy~2:
	Each raw value $x_{i}$  is converted into integral value $s_{i}$ at each point $i$ as follows: $s_{i} = x_{i} (t_{i} - t_{i-1}) +  s_{i-1}$. 
	Then all time series are re-sampled such that each re-sampled value $x_{j}$ at every $n$ (the lowest sampling rate) steps is calculated as follows: $x_{j} = \frac{s_{i}-s_{i-n}}{t_{i} - t_{i-n}}$. The time $t_{i}$ (of value $s_{i}$) is the time that is after the corresponding time to $x_{j}$. 
\end{itemize}

\section{Datasets description}
\label{sec:datasets}

\begin{figure*}[t!]
	\centering
	\begin{subfigure}{.45\textwidth}
		\centering
		
		\begin{tikzpicture}[{black, circle, draw, inner sep=0}]
		\tikzset{nodes={draw,rounded corners},minimum height=0.9cm,minimum width=0.9cm, font=\footnotesize}	
		\tikzset{anomalous/.append style={fill=easyorange}}
		\tikzset{rc/.append style={fill=easyorange}}
		
		\node (P) at (3,1.5) {Pub} ;
		\node (CPU) at (0,0) {Cpu};
		\node (RAM) at (1.5,0) {Ram};
		\node (DiskR) at (6,0) {DiskR};
		\node (DiskW) at (4.5,0) {DiskW};
		\node (M) at (3,0) {Mes};
		\node (CON) at (3,-1.5) {Con};
		
		\draw[->,>=latex] (P) -- (CPU);
		\draw[->,>=latex] (P) -- (M);
		\draw[->,>=latex] (CON) -- (CPU);
		\draw[->,>=latex] (CON) -- (M);
		\draw[->,>=latex] (P) -- (RAM);
		\draw[->,>=latex] (CON) -- (RAM);
		\draw[->,>=latex] (P) -- (DiskW);
		\draw[->,>=latex] (CON) -- (DiskW);
		\draw[->,>=latex] (P) -- (DiskR);
		\draw[->,>=latex] (CON) -- (DiskR);
		
		\draw[->,>=latex] (P) to [out=180,in=135, looseness=2] (P);
		\draw[->,>=latex] (CPU) to [out=180,in=135, looseness=2] (CPU);
		\draw[->,>=latex] (M) to [out=180,in=135, looseness=2] (M);
		\draw[->,>=latex] (RAM) to [out=180,in=135, looseness=2] (RAM);
		\draw[->,>=latex] (DiskW) to [out=0,in=45, looseness=2] (DiskW);
		\draw[->,>=latex] (DiskR) to [out=0,in=45, looseness=2] (DiskR);
		\draw[->,>=latex] (CON) to [out=-30,in=15, looseness=2] (CON);
		
		\end{tikzpicture}
		\caption{MoM system.}
		\label{fig:controlled_graph}
		
	\end{subfigure}\hfill
	\begin{subfigure}{.45\textwidth}
		\centering
		\begin{tikzpicture}[{black, circle, draw, inner sep=0}]
		\tikzset{nodes={draw,rounded corners},minimum height=0.9cm,minimum width=0.9cm, font=\footnotesize}	
		\tikzset{anomalous/.append style={fill=easyorange}}
		\tikzset{rc/.append style={fill=easyorange}}
		
		\node (PMDB) at (0.5,0) {PMDB} ;
		\node (MDB) at (2,0) {MDB};
		\node (CMB) at (3.5, 0.75) {CMB};
		\node (MB) at (5,1.5) {MB};
		\node (LMB) at (6.5, 0.75) {LMB};
		\node (RTMB) at (3.5, -0.75) {RTMB} ;
		\node (GSIB) at (5, -1.5) {GSIB};
		\node (ESB) at (6.5, -0.75) {ESB} ;
		
		\draw[->,>=latex] (PMDB) -- (MDB);
		\draw[->,>=latex] (MDB) -- (CMB);
		\draw[->,>=latex] (CMB) -- (MB);
		\draw[->,>=latex] (CMB) -- (RTMB);
		\draw[->,>=latex] (MB) -- (LMB);
		\draw[->,>=latex] (MDB) -- (RTMB);
		\draw[->,>=latex] (RTMB) -- (GSIB);
		\draw[->,>=latex] (RTMB) -- (ESB);
		\draw[->,>=latex] (GSIB) -- (ESB);
		
		\draw[->,>=latex] (PMDB) to [out=90,in=45, looseness=2] (PMDB);
		\draw[->,>=latex] (MDB) to [out=90,in=45, looseness=2] (MDB);
		\draw[->,>=latex] (CMB) to [out=90,in=45, looseness=2] (CMB);
		\draw[->,>=latex] (MB) to [out=0,in=45, looseness=2] (MB);
		\draw[->,>=latex] (LMB) to [out=0,in=45, looseness=2] (LMB);
		\draw[->,>=latex] (RTMB) to [out=-90,in=-45, looseness=2] (RTMB);
		\draw[->,>=latex] (GSIB) to [out=-45,in=0, looseness=2] (GSIB);
		\draw[->,>=latex] (ESB) to [out=0,in=45, looseness=2] (ESB);	
		
		\end{tikzpicture}
		\caption{Ingestion IT monitoring system.}
		\label{fig:storm_graph}
	\end{subfigure}
	
	\begin{subfigure}{.45\textwidth}
		\centering
		\begin{tikzpicture}[{black, circle, draw, inner sep=0}]
		\tikzset{nodes={draw,rounded corners},minimum height=0.9cm,minimum width=0.9cm, font=\footnotesize}
		
		\node (CpuG) at (3,0.2) {CpuG};
		\node (CpuHttp) at ({2.5*cos(36)},{2.5*sin(36)}) {CpuH};
		\node (NbProcessHttp) at ({2.5*cos(76)},{2.5*sin(76)}) {NPH};
		\node (RamHttp) at ({3*cos(116)},{2.5*sin(116)}) {RamH};
		\node (NetOutG) at ({3*cos(208)},-{0.5}) {NetOut};
		\node (NCM) at ({2.5*cos(248)},-{0.8}) {NCM};
		\node (NbProcessPhp) at (-1, 0.8) {NPP};
		\node (CpuPhp) at (1, 0.5) {CpuP};
		\node (DiskW) at (1.2,-{0.8}) {DiskW};
		\node (NetInG) at (-3, 0.8) {NetIn};

		\draw[->,>=latex] (NetInG) -- (NbProcessHttp);
		\draw[->,>=latex] (NetInG) -- (NetOutG);
		\draw[->,>=latex] (NetInG) -- (NCM);
		\draw[->,>=latex] (NbProcessHttp) -- (NbProcessPhp);
		\draw[->,>=latex] (NbProcessHttp) -- (CpuHttp);
		\draw[->,>=latex] (NbProcessHttp) -- (RamHttp);
		\draw[->,>=latex] (NbProcessPhp) -- (NCM);
		\draw[->,>=latex] (NbProcessPhp) -- (CpuPhp);
		\draw[->,>=latex] (NCM) -- (DiskW);
		\draw[->,>=latex] (NCM) -- (NetOutG);
		\draw[->,>=latex] (CpuHttp) -- (CpuG);
		\draw[->,>=latex] (CpuPhp) -- (CpuG);
		\draw[->,>=latex] (NCM) -- (CpuG);
		\draw[->,>=latex] (DiskW) -- (CpuG);
		
		\draw[->,>=latex] (NetInG) to [out=180,in=135, looseness=2] (NetInG);
		\draw[->,>=latex] (NbProcessHttp) to [out=90,in=45, looseness=2] (NbProcessHttp);
		\draw[->,>=latex] (NbProcessPhp) to [out=180,in=135, looseness=2] (NbProcessPhp);
		\draw[->,>=latex] (CpuHttp) to [out=90,in=45, looseness=2] (CpuHttp);
		\draw[->,>=latex] (CpuPhp) to [out=90,in=45, looseness=2] (CpuPhp);
		\draw[->,>=latex] (RamHttp) to [out=180,in=135, looseness=2] (RamHttp);
		\draw[->,>=latex] (NetOutG) to [out=180,in=135, looseness=2] (NetOutG);
		\draw[->,>=latex] (CpuG) to [out=0,in=45, looseness=2] (CpuG);
		\draw[->,>=latex] (NCM) to [out=280,in=235, looseness=2] (NCM);
		\draw[->,>=latex] (DiskW) to [out=280,in=235, looseness=2] (DiskW);
		\end{tikzpicture} 
		\caption{Web-Activity.}
		\label{fig:Webactivity}
	\end{subfigure}
	\hfill 
	\begin{subfigure}{.45\textwidth}
		\centering
		\begin{tikzpicture}[{black, circle, draw, inner sep=0}]
		\tikzset{nodes={draw,rounded corners},minimum height=0.9cm,minimum width=0.9cm, font=\footnotesize}
		
		\node (MUsaP) at (3,0) {MUP};
		\node (MUseP) at (1.5,0) {MUGP};
		\node (CpuUsaP) at (3,3) {CUP};
		\node (CpuUseP) at (1.5,3) {CUGP};
		\node (MUsaV) at (-3,0) {MUV};
		\node (MUseV) at (-1.5,0) {MUGV};
		\node (CpuUsaV) at (-3, 3) {CUV};
		\node (CpuUseV) at (-1.5, 3) {CUGV};
		\node (RP) at (3,1.5) {RP};
		\node (RV) at (-3, 1.5) {RV};
		\node (ChP) at (0,3) {ChP};
		\node (ChIE) at (0, 0) {ChIE};
		\node (T) at (0, 1.5) {T};

		\draw[->,>=latex] (MUsaP) -- (MUseP);
		\draw[->,>=latex] (CpuUsaP) -- (CpuUseP);
		\draw[->,>=latex] (MUseP) -- (RP);
		\draw[->,>=latex] (CpuUseP) -- (RP);
		\draw[->,>=latex] (MUsaV) -- (MUseV);
		\draw[->,>=latex] (CpuUsaV) -- (CpuUseV);
		\draw[->,>=latex] (MUseV) -- (RV);
		\draw[->,>=latex] (CpuUseV) -- (RV);
		\draw[->,>=latex] (MUseP) -- (ChP);
		\draw[->,>=latex] (CpuUseP) -- (ChP);
		\draw[->,>=latex] (MUseV) -- (ChIE);
		\draw[->,>=latex] (CpuUseV) -- (ChIE);
		\draw[->,>=latex] (ChP) -- (T);
		\draw[->,>=latex] (ChIE) -- (T);
		\draw[->,>=latex] (RP) -- (ChP);
		\draw[->,>=latex] (RV) -- (ChIE);

		\draw[->,>=latex] (ChP) to [out=90,in=45, looseness=2] (ChP);
		\draw[->,>=latex] (CpuUseV) to [out=90,in=45, looseness=2] (CpuUseV);
		\draw[->,>=latex] (CpuUseP) to [out=90,in=45, looseness=2] (CpuUseP);
		\draw[->,>=latex] (RP) to [out=0,in=45, looseness=2] (RP);
		\draw[->,>=latex] (CpuUsaP) to [out=0,in=45, looseness=2] (CpuUsaP);
		\draw[->,>=latex] (MUsaP) to [out=0,in=45, looseness=2] (MUsaP);
		\draw[->,>=latex] (RV) to [out=180,in=135, looseness=2] (RV);
		\draw[->,>=latex] (CpuUsaV) to [out=180,in=135, looseness=2] (CpuUsaV);
		\draw[->,>=latex] (MUsaV) to [out=180,in=135, looseness=2] (MUsaV);
		\draw[->,>=latex] (ChIE) to [out=270,in=315, looseness=2] (ChIE);
		\draw[->,>=latex] (MUseV) to [out=270,in=315, looseness=2] (MUseV);
		\draw[->,>=latex] (MUseP) to [out=270,in=315, looseness=2] (MUseP);
		\draw[->,>=latex] (T) to [out=0,in=45, looseness=2] (T);
		\end{tikzpicture} 
		\caption{Antivirus-Activity.}
		\label{fig:Antivirusactivity}
	\end{subfigure}
	
	\caption{Summary causal graphs for different datasets:  MoM system based on Publish/Subscribe architecture (a), Ingestion IT monitoring system (b), Web-Activity (c) and Antivirus-Activity (d). Those summary causal graphs are constructed either by IT monitoring system experts or directly using the system topology.}
	\label{fig:datasets}
\end{figure*}
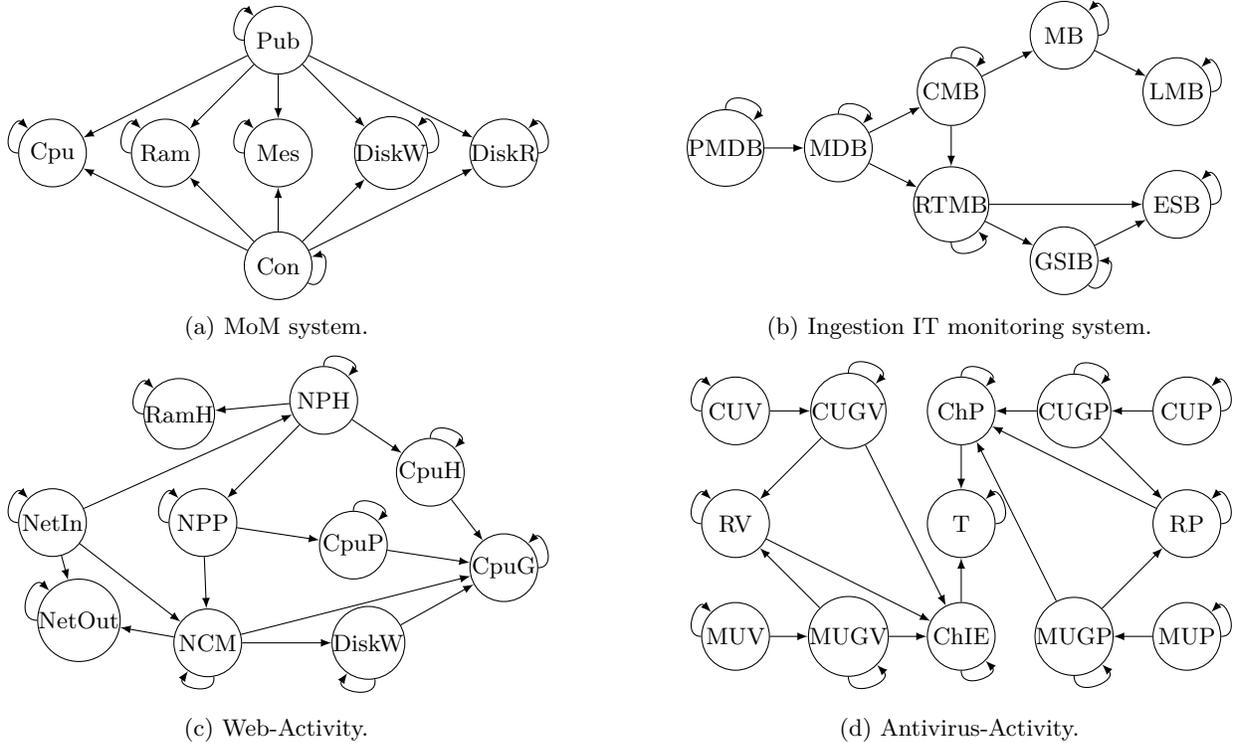%

In this section, we present the summary causal graph (the window causal graph and the extended summary causal graph are not available) and the datasets\footnote{All datasets are available at \url{https://easyvista2015-my.sharepoint.com/:f:/g/personal/aait-bachir_easyvista_com/ElLiNpfCkO1JgglQcrBPP9IBxBXzaINrM5f0ILz6wbgoEQ?e=OBTsUY}} for each case study. All summary causal graphs are constructed either by IT monitoring experts or directly using the system topology.
Note that all data points are collected using Nagios\footnote{\url{https://www.nagios.org/}}, an open-source software that monitors systems, networks and infrastructure, and which gives the timestamp according to the collection time which does not necessarily correspond to the real time of the value.
In addition, on some of the case studies, the alignment between time series is not guaranteed as data collection is performed by different plugins with different starting times and different sampling rates \citep{Holzinger_2021}.
In Table~\ref{tab:summary_algo}, we also classify datasets with respect to the different assumptions needed by causal discovery algorithms and to other different characteristics.
In Appendix, we also give additional information on the datasets.

\subsection{MoM activity datasets}
\label{sec:datasets_controlled_data}
First, we consider two Middleware oriented Message datasets which we denote as MoM 1 and MoM 2 and which are defined through the monitoring of an IT pipeline which ingests incoming messages based on a Publish/Subscribe architecture. These datasets contain seven different time series such as MoM 1 consists of $288$ timestamps, MoM 2 consists of $364$ timestamps, and they are both collected with a one-second sampling rate. Note that there is no overlapping time between these datasets. 
The corresponding summary causal graph is presented in Figure~\ref{fig:controlled_graph} where \textit{Pub} represents the publish rate that monitor the number of messages per seconds;  \textit{Con} is the number of consumers; \textit{Mes} represents the number of messages remaining in the queue; \textit{Cpu} represents the percentage of used CPU; \textit{Ram} represents the percentage of used RAM; \textit{DiskW} represents the Disk write in Kbytes/second; \textit{DiskR} represents the Disk read in Kbytes/second. 
There might exist additional links between \textit{Cpu, Ram,  DiskW} and \textit{DiskR} under extreme conditions and abnormal behavior but these relations are excluded from the graph since there is no incident or clear anomaly that is detected for these datasets.
Note that, for these datasets, timestamps of different time series are aligned.



\subsection{Ingestion activity dataset}
\label{sec:datasets_storm}
We also consider a dataset introduced in \cite{Assaad_2023} which we denote as the Ingestion dataset. This dataset contains eight time series which consist of $991$ timestamps collected with a one-minute sampling rate. The corresponding summary causal graph which is constructed using Storm ingestion topology that describes the relations between the inputs and outputs of each Bolt is provided in Figure~\ref{fig:storm_graph} where \textit{PMDB} represents the extraction of some information about the messages received by the Storm ingestion system; \textit{MDB} refers to an activity of a process that orients messages to other processes with respect to different types of messages; \textit{CMB} represents the activity of extraction of metrics from messages; MB represents the activity of insertion of data in a database; \textit{LMB} reflects the updates of the last values of metrics in Cassandra; \textit{RTMB} represents the activity of searching to merge data with information coming from the check message bolt; \textit{GSIB} represents the activity of insertion of historical status in database; \textit{ESB} represents the activity of writing data in Elasticsearch. All values are calculated by multiplying the number of messages executed on the specific bolt in a given time window by the average execution latency in the same time window, and then dividing it by the time window which corresponds to $10$ minutes.
Note that, for this dataset, timestamps of different time series are aligned. 

\subsection{Web activity dataset}
\label{sec:datasets_web_activity}

We consider a dataset that reflects the activity in a web server.
This dataset contains ten time series collected with a one-minute sampling rate. 
The raw data of this case study were initially misaligned. In  order to align them, we use the two pre-processing strategies described in Section~\ref{sec:pre-process}.
We denote the dataset pre-processed using Strategy 1 as Web 1 and the dataset pre-processed using Strategy 2 as Web 2.
The two processed datasets contains $3000$ timestamps.
The corresponding summary causal graph is presented in Figure~\ref{fig:Webactivity} where \textit{NetIn} represents the data received by the network interface card in Kbytes/second; 
\textit{NetOut} represents the data transmitted out by the network interface card in Kbytes/second;
\textit{NPH} represents the number of HTTP processes; 
\textit{NPP} represents the number of PHP processes; 
\textit{NCM} represents the number of open MySql connections which are started by PHP processes;
\textit{CpuH} represents the percentage of CPU used by all HTTP processes; 
\textit{RamH} represents the percentage of RAM used by all HTTP processes; 
\textit{CpuP} represents the percentage of CPU used by all PHP processes;
\textit{DiskW represents the Disk write in Kbytes/second};
\textit{CpuG} represents the percentage of global CPU usage.  

\subsection{Antivirus activity dataset}
\label{sec:datasets_antivirus_activity}

Lastly, we consider a  dataset which depicts the impacts of antivirus activity in servers.
This dataset contains 13 time series such that $3$ of them are collected with a one-minute sampling rate and the rest with a five-minutes sampling rate.
The raw data of this case study were initially misaligned. In  order to align them, we use the two pre-processing strategies described in Section~\ref{sec:pre-process}, leading to the dataset Antivirus 1 for Strategy 1 and Antivirus 2 for Strategy 2. 
The two processed datasets consist of $1321$ timestamps.
The corresponding summary causal graph is presented in Figure~\ref{fig:Antivirusactivity} where \textit{CUV} represents the percentage of CPU usage of antivirus processes in server V; 
\textit{CUGV} represents the percentage of CPU usage of the global server V;
\textit{MUV} represents the percentage of memory usage of antivirus process; \textit{MUGV} represents the percentage  of global memory usage of the server; 
\textit{RV} represents the Disk IO read in Kbytes/second; 
\textit{ChIE} refers to the required duration in seconds to open an \textit{IE browser} on server V;
\textit{CUP} represents the percentage of CPU usage of antivirus processes in server P;
\textit{CUGP} represents the percentage of CPU usage of the global server P; 
\textit{MUP} represents the percentage of memory usage of antivirus process; 
\textit{MUGP} represents the percentage of global memory usage of the server; 
\textit{RP} represents the Disk IO read in Kbytes/second; 
\textit{ChP} represents refers to the required duration in seconds to open a \textit{ CITRIX Portal} on server P;
\textit{T} represents the global time in seconds required to open a CITRIX portal and open the IE browser.

\section{Results}
\label{sec:results}

\begin{table*}[t]
	\caption{Results for real IT monitoring datasets where $\gamma_{max}$ is set according to the $15$ seconds delay rule for MoM datasets and to the $15$ minutes delay rule for Ingestion, Web and Antivirus datasets.  We report the F1-score.} \label{tab:results_real_gm_15_1}
	\centering
	\begin{tabular}{r|c|c|c|c|c|c|c}
		\hline 
		& MoM 1 & MoM 2 &  Ingestion & Web 1 & Web 2 & Antivirus 1 & Antivirus 2\\ \hline
		GCMVL& $0.0$ & $0.0$ & $0.2$  & $0.2$  & $0.0$    &$0.08$ & $0.0$\\
		Dynotears& $0.26$ & $0.2$ & $0.14$&  $0.23$  & $0.3$     &$0.18$ & $0.19$\\
		PCMCI$^+$ & $\textbf{0.4}$ & $0.0$  & $0.0$ & $0.23$  & $0.3$     &$0.04$ & $0.11$\\
		PCGCE & $0.0$ & $0.12$ & $0.12$ &  $0.22$
		&   $0.15$      &$0.3$ & $\textbf{0.45}$\\
		VLiNGAM& $0.0$ & $0.0$ & $0.19$ & $\textbf{0.29}$
		& $0.18$     &$0.15$ & $0.22$\\
		TiMINo& $0.0$ & $0.17$ & $0.18$ & $0.0$ &  $0.0$     &$0.0$ & $0.0$\\
		NBCB-w& $\textbf{0.4}$ & $0.0$ & $0.13$&  $0.23$
		&    $0.3$     &$0.14$ & $0.24$\\
		NBCB-e& $0.13$ & $\textbf{0.29}$ & $\textbf{0.27}$&  $0.19$
		& $\textbf{0.42}$     &$\textbf{0.31}$ & $\textbf{0.45}$\\
		CBNB-w& $\textbf{0.4}$ & $0.0$ & $0.15$&  $0.23$
		&    $0.3$     &$0.17$ & $0.16$\\
		CBNB-e& $0$ & $0.24$ & $0.13$&  $0.22$
		& $0.29$     &$\textbf{0.31}$ & $0.38$
	\end{tabular} 
\end{table*}

In this section we evaluate the performance of each causal discovery algorithm presented in Section~\ref{sec:setup} on each dataset presented in Section~\ref{sec:datasets}. Since we only have access to the true summary causal graph, we evaluate the detection of oriented edges of the summary causal graph (for algorithms that detect a window causal graphs or an extended summary causal graph, we start by inferring the window causal graphs or the extended summary causal graph then deduce the summary causal graph from it) using the F1-score.
Recall that as mentioned in Section~\ref{sec:setup}, $\gamma_{max}$ is set according to the $15$ seconds delay rule for MoM datasets and the $15$ minutes delay rule for the other datasets. 

All results~\footnote{The code is available at \url{https://github.com/ckassaad/Case_Studies_of_Causal_Discovery_from_IT_Monitoring_Time_Series}} are presented in Table~\ref{tab:results_real_gm_15_1}.  GCMVL exhibits poor performance on MoM 1, MoM 2, Web 2, Antivirus 1, and Antivirus 2 datasets. However, it shows better performance on Ingestion and Web 1 datasets, achieving at most an F1-score of $0.2$. Dynotears demonstrates relatively better performance compared to GCMVL across all datasets, except for the Ingestion dataset. 
PCMCI$^+$, NBCB-w and CBNB-w achieve the highest F1-score of $0.4$ on the MoM 1 dataset along with NBCB-w and CBNB-w. However, PCMCI$^+$ performs poorly on the MoM 2 dataset and on the Ingestion dataset. On Web 1 and Web 2, it respectively achieves better F1-scores of $0.23$ and $0.3$, and lower F1-scores of $0.04$ and $0.11$ on Antivirus 1 and Antivirus 2. 
PCGCE exhibits low performance on the MoM 1 dataset but has better performance on MoM 2, Ingestion, Web 1, and Web 2 with F1-scores of $0.12$, $0.12$, $0.22$, and $0.15$, respectively. Remarkably, PCGCE performs well on Antivirus datasets, achieving the highest F1-score of $0.45$ for Antivirus 2, along with NBCB-e. 
VLiNGAM shows poor performance on MoM 1 and MoM 2 datasets. However, it shows better performance on the other datasets and achieves the highest F1-score of $0.29$ on Web 1 dataset. TiMINo performs poorly on the majority of datasets but shows better performance on MoM 2 and Ingestion datasets. 
For all datasets, NBCB-w and CBNB-w (which are two hybrids methods which comines PCMCI$^+$ with VLiNGAM) either outperform PCMCI$^+$ or performs equally to PCMCI$^+$ and they outperform VLiNGAM for only 3 datasets.
On the other hand, for most datasets, NBCB-e and CBNB-e (which are two hybrids methods which comines PCGCE with VLiNGAM) outperform PCCGE and VLiNGAM except for Web 1 and Antivirus 2.
In addition, in most datasets NBCB-e outperform CBNB-e.
Notably, NBCB-e achieves the best F1-scores in most datasets (MoM 2, Ingestion, Web 2, Antivirus 1 and Antivirus 2) and has relatively good performance on the other datasets.

Note that all algorithms, except GCMVL and TiMINo, have better performance on the Antivirus dataset when pre-processing Strategy~2 is applied.

Overall, according to the performance of each algorithm, NBCB-e seems to be the best choice across all datasets. However, it is important to note that the best performance achieved ($0.45$ in Antivirus 2) is far from being satisfactory.



\section{Discussion}
\label{sec:elaboration} 

As shown in the previous section, the results of causal discovery algorithms considered in this work are not satisfactory. Most probably this is due to the violation of the assumptions that these algorithms rely on.  
These algorithms typically assume \textit{Consistency throughout time} and \textit{Stationarity}, yet IT systems exhibit different states (e.g., normal, warning, critical). Transitions between these states can potentially induce changes in the causal strengths between metrics or even completely alter the underlying causal graph. 
Consequently, it will be interesting to test methods that relax this assumption such as  CD-NOD \citep{huang2020causal}, R-PCMCI \citep{saggioro2020reconstructing}, LoSST \citep{kummerfeld2013tracking}, and SDCI \citep{rodas2021causal}. 
On top of that, the linearity assumption is not verified in any of the datasets, thus it would be interesting to test nonlinear methods, for example, by using non-linear independence test in constraint-based methods and non-linear regression models in semi-parametric methods. 

It is also assumed throughout this paper that the full time graph is acyclic which coincides well with the summary causal graphs of our case studies. However, in general, considering low sampling rate challenges the legitimacy of this assumption as there might be two lagged causal relation with a lag smaller than the time delay between each two collected data points. 
Additionally, in this work we only focused on continuous data, however, IT systems comprise not only continuous variables but also ordinal variables ($e.g.$, CPU frequency) and nominal variables ($e.g.$, device states: normal, busy, overcharged) which can help improve performance. To address mixed data types, it is worth exploring independent measures and tests for mixed data. Prominent methods include SCPC \citep{cui2016copula}, MGVI \citep{tsagris2018constraint}, MIIC \citep{cabeli2020learning}, LH \citep{marx2021estimating}, RAVK \citep{rahimzamani2018estimators}, MS \citep{mesner2020conditional}, and CMIh \citep{e24091234}, as they can be integrated directly into constraint-based  algorithms. 
Furthermore, missing data  poses a significant challenge in determining the causal graph. This issue could be fixed by methods like MVPC \citep{tu2019causal} and CBR-PC \citep{gain2018structure}.  
Finally, there might always be hidden common causes in the system, so using methods based on FCI~\citep{Spirtes_2000,Gerhardus_2020,Assaad_2022b} might be useful, but in this case, the graph will be in most cases less informative and less interpretable by IT monitoring experts.  





\section{Conclusion}
\label{sec:conclusion}

IT systems are crucial for the success of modern businesses, and monitoring them is essential to ensure their proper functioning. Causal discovery techniques offer powerful tools for identifying the root causes of issues, optimizing IT systems, and predicting future problems. However, the analysis of IT monitoring data presents challenges due to its complexity and volume. The case study presented in this paper shows both the potential benefits and ongoing challenges of applying causal discovery algorithms to IT monitoring data. This area should continue to be an active field of research and development, with the aim of improving the efficiency and performance of IT systems in diverse industries and applications.

%

\bibliographystyle{plain}
\bibliography{Ref}

\appendix
\newpage
\newpage

\section{Appendix}

In the following, we start by presented additional experimental results then present an examination of the datasets we have considered.

\subsection{Additional results}

\begin{table*}[h]
	\caption{Results for real IT monitoring datasets for $\gamma_{max}=15$.  We report the F1-score. } \label{tab:results_real_gm_15_2}
	\centering
	\begin{tabular}{r|c|c|c|c|c|c|c}
		\hline 
		& MoM 1 & MoM 2 &  Ingestion & Web 1 & Web 2 & Antivirus 1 & Antivirus 2\\ \hline
		GCMVL& $0.0$ & $0.0$ & $0.2$  & $0.29$ & $0.0$    &$0.08$ & $0.0$\\
        Dynotears& $0.26$ & $0.2$ & $0.14$& $0.24$ & $\textbf{0.34}$     &$0.19$ & $0.25$\\
		PCMCI$^+$ & $\textbf{0.4}$ & $0.0$  & $0.0$ &  $0.22$ & $0.31$     &$0.1$ & $0.13$\\
		PCGCE & $0.0$ & $0.12$ & $0.12$ &  $\textbf{0.3}$ &   $0.27$      &$0.27$ & $\textbf{0.26}$\\
		VLiNGAM& $0.0$ & $0.0$ & $0.19$ &  $0.24$  & $0.17$     &$0.19$ & $0.16$\\
  		TiMINo& $0.0$ & $0.17$ & $0.18$ & $0.0$ &  $0.13$     &$0.0$ & $0.07$\\
  		NBCB-w& $\textbf{0.4}$ & $0.0$ & $0.13$&  $0.18$
 &    $0.23$     &$0.13$ & $0.19$\\
            NBCB-e& $0.13$ & $\textbf{0.29}$ & $\textbf{0.27}$& $0.19$ & $0.22$     &$0.22$ & $0.15$\\
            CBNB-w& $\textbf{0.4}$ & $0.0$ & $0.15$&  $0.22$ &  $0.29$ &$0.2$ & $0.19$\\
            CBNB-e& $0.0$ & $0.24$ & $0.13$&  $0.23$ &  $0.33$ &$\textbf{0.28}$ & $0.22$
	\end{tabular} 
\end{table*}

\begin{table*}[h]
	\caption{Results for real IT monitoring datasets for $\gamma_{max}=10$.  We report the F1-score. } \label{tab:results_real_gm_10}
	\centering
	\begin{tabular}{r|c|c|c|c|c|c|c}
		\hline 
		& MoM 1 & MoM 2 & Ingestion & Web 1 & Web 2 & Antivirus 1 & Antivirus 2\\ \hline
        GCMVL& $0.0$ & $0.0 $  & $0.0$ & $\textbf{0.32}$  & $0.0$ & $0.09$ & $0.0$\\
        Dynotears& $\textbf{0.36}$ & $0.14$& $0.14$  & $0.22$ & $\textbf{0.39}$ & $0.18$ & $0.22$ \\ 
        PCMCI$^+$ & $0.0$ & $0.0$ & $0.0$ & $0.22$  & $0.31$ & $0.07$ & $0.14$\\
        PCGCE & $0.0$ & $0.0$ & $0.11$ & $0.27$  & $0.24$ &$ \textbf{0.33}$ & $0.27$\\
        VLiNGAM& $0.27$ & $0.09$ & $\textbf{0.27}$ & $0.22$ & $0.18$ & $0.19$ & $0.16$\\
        TiMINo& $0.0$ & $0.17$ & $0.17$ & $0.0$  & $0.0$ & $0.06$ &  $0.06$\\
        NBCB-w& $0.15$ & $0.0$& $0.13$  & $0.19$ & $0.23$ & $0.15$& $0.25$\\
        NBCB-e& $0.13$ & $\textbf{0.2}$& $0.18$ & $0.25$  & $0.22$ & $0.26$ & $0.21$\\
        CBNB-w& $0.15$ & $0.0$ & $0.16$&  $0.22$ &  $0.29$ &$0.2$ & $0.21$\\
        CBNB-e& $0.0$ & $0.12$ & $0.11$&  $0.21$ &  $0.26$ &$\textbf{0.33}$ & $\textbf{0.29}$

	\end{tabular} 
\end{table*}

\begin{table*}[h]
	\caption{Results for real IT monitoring datasets for $\gamma_{max}=5$.  We report the F1-score.} \label{tab:results_real_gm_5}
	\centering
	\begin{tabular}{r|c|c|c|c|c|c|c}
		\hline 
		& MoM 1 & MoM 2 & Ingestion & Web 1  & Web 2 & Antivirus 1 & Antivirus 2\\ \hline
		GCMVL& $0.0$ & $0.0$  & $0.0$  & $0.19$    & $0.0$ & $0.08$ & $0.0$\\
        Dynotears& $0.27$ & $\textbf{0.21}$& $0.14$  & $0.22$    & $0.3$ & $0.18$ & $0.17$\\
		PCMCI$^+$ & $0.0$  & $0.15$ & $0.0$  & $0.17$    & $0.32$ & $0.04$ & $0.11$\\
		PCGCE & $\textbf{0.31}$ & $0.0$ & $0.22$   & $0.21$ & $0.34$ & $0.3$ & $0.36$\\
		VLiNGAM& $0.0$ & $0.19$ & $\textbf{0.25}$  & $0.23$  & $0.2$ & $0.18$ & $0.18$\\
  		TiMINo& $0.0$ & $0.0$ & $0.18$  & $0.0$  & $0.0$  & $0.0$ & $0.0$\\
		NBCB-w& $0.0$ & $0.12$& $0.13$  & $0.2$ & $0.23$ & $0.13$ & $0.3$\\
		NBCB-e& $0.27$ & $0.0$& $0.11$  & $\textbf{0.24}$  & $\textbf{0.42}$ & $0.29$ & $\textbf{0.38}$\\
        CBNB-w& $0.0$ & $0.13$ & $0.15$&  $\textbf{0.24}$ &  $0.29$ &$0.18$ & $0.18$\\
        CBNB-e& $\textbf{0.31}$ & $0.0$ & $0.13$&  $0.15$ &  $0.38$ &$\textbf{0.33}$ & $0.27$
	\end{tabular}
\end{table*}

\begin{table*}[t]
	\caption{Results for real IT monitoring datasets for $\gamma_{max}=3$.  We report the F1-score. } \label{tab:results_real_gm_3}
	\centering
	\begin{tabular}{r|c|c|c|c|c|c|c}
		\hline 
		& MoM 1 & MoM 2 &  Ingestion & Web 1 & Web 2 & Antivirus 1 & Antivirus 2\\ \hline
		GCMVL& $0.0$ & $0.0$ & $0.14$  & $0.2$  & $0.0$    &$0.08$ & $0.0$\\
        Dynotears& $0.14$ & $\textbf{0.3}$ & $0.14$&  $0.23$  & $0.3$     &$0.18$ & $0.19$\\
		PCMCI$^+$ & $0.0$ & $0.0$  & $0.0$ & $0.23$  & $0.3$     &$0.04$ & $0.11$\\
		PCGCE & $\textbf{0.15}$ & $0.0$ & $0.22$ &  $0.22$
 &   $0.15$      & $0.3$ & $\textbf{0.45}$\\
		VLiNGAM& $0.0$ & $0.0$ & $\textbf{0.38}$ & $\textbf{0.29}$
  & $0.18$     &$0.15$ & $0.22$\\
  		TiMINo& $0.0$ & $0.17$ & $0.18$ & $0.0$ &  $0.0$     &$0.0$ & $0.0$\\
  		NBCB-w& $0.0$ & $0.0$ & $0.15$&  $0.23$
 &    $0.3$     &$0.14$ & $0.24$\\
            NBCB-e& $0.14$ & $0.0$ & $0.22$&  $0.19$
   & $\textbf{0.42}$     &$\textbf{0.31}$ & $\textbf{0.45}$\\
       CBNB-w& $0.0$ & $0.0$ & $0.16$&  $0.23$ &  $0.3$ &$0.17$ & $0.16$\\
        CBNB-e& $0.0$ & $0.0$ & $0.13$&  $0.22$ &  $0.29$ &$\textbf{0.31}$ & $0.38$
	\end{tabular} 
\end{table*}

Tables \ref{tab:results_real_gm_15_2}, \ref{tab:results_real_gm_10}, \ref{tab:results_real_gm_5}, and \ref{tab:results_real_gm_3} present the F1-scores for each method using different values of $\gamma_{max}$ (15, 10, 5, and 3, respectively). Among these methods, GCMVL performs poorly on all datasets, except for Web 1 dataset where it achieves the highest F1-scores of $0.32$, when $\gamma_{max} =$ $10$.  Dynotears demonstrates stable performance across various datasets when $\gamma_{max}$ is varied. It achieves the highest F1-scores on the Web 2 dataset with a large values of $\gamma_{max}$, and on the MoM 2 dataset with a small values of $\gamma_{max}$.

PCMCI$^{+}$ exhibits poor performance on the MoM, Ingestion, and Antivirus datasets, except for the MoM 1 dataset when $\gamma_{max}$ is set to $15$, where it achieves an F1-score of $0.4$. However, PCMCI$^{+}$ shows better performance on the Web datasets. 
PCGCE achieves the highest F1-score on the MoM 1 dataset when $\gamma_{max}=3$ and $\gamma_{max}=5$ however  it F1-score drops to zero for $\gamma_{max}=10$ and $15$. For the MoM 2 dataset, PCGCE has almost always a zero F1-score and for the Ingestion dataset it has relatively a low performance.
However, when it comes to the Web and Antivirus datasets, PCGCE consistently exhibits good performance across all values of $\gamma_{max}$. 
VLiNGAM consistently achieves high F1-scores on the Ingestion dataset for all values of $\gamma_{max}$ except when $\gamma_{max}=3$, and it performs better on the Web and Antivirus datasets compared to the MoM datasets. 
TiMINo performs poorly on the majority of the datasets, but it demonstrates stable performance on the Ingestion dataset regardless of the value of $\gamma_{max}$. It shows a similar conclusion on the MoM 2 dataset, except when $\gamma_{max}$ is set to $5$.
NBCB-w and CBNB-w achieve the best F1-score of $0.4$ on the MoM 1 dataset when $\gamma_{max}$ is set to $15$, and it generally performs better on the Web and Antivirus datasets compared to the other datasets. 
NBCB-e tends to achieve the highest F1-scores in most cases. It achieves the highest F1-scores on the MoM 2 and Ingestion datasets when $\gamma_{max}$ is large, and on the Web and Antivirus datasets when $\gamma_{max}$ is smaller, meanwhile, it should be noted that as $\gamma_{max}$ increases, its performance remains comparative on these datasets.
Similarly, CBNB-e has the best F1-score in Antivirus 1 when $\gamma_{max}$ is set to $10$ and $15$ and in Antivirus 2 when $\gamma_{max}$ is set to $10$.  

In summary, it appears that there is no single method that works well for all datasets. If the value of $\gamma_{max}$ is unknown, NBCB-e and PCGCE are the recommended choice for the Antivirus datasets, as they consistently performs well across these datasets for all values of $\gamma_{max}$. Similarly, NBCB-e, PCGCE and PCMCI$^+$ are the recommended choice for the Web datasets (Dynotears was excluded because as shown in Figures~\ref{fig:web_infer_graph1} and \ref{fig:web_infer_graph2}, it gives almost a fully connected graph for the Web datasets).
For the Ingestion dataset, VLiNGAM is the best choice. Lastly, Dynotears is a better option for the MoM 1 and MoM 2 datasets due to its stability across different values of $\gamma_{max}$. 

However, it is important to note that the best performance achieved ($0.45$ in Antivirus 2) is far from being satisfactory for real world application.



%

In Figures~\ref{fig:controlled_infer_graph_1},\ref{fig:controlled_infer_graph_2},\ref{fig:storm_infer_graph}, \ref{fig:web_infer_graph1}, \ref{fig:web_infer_graph2}, \ref{fig:antivirus_infer_graph1} and \ref{fig:antivirus_infer_graph2} we also give the the inferred graphs that correspond to the results in Table~\ref{tab:results_real_gm_15_1} (where $\gamma_{max}$ is set using the 15 seconds rule for the MoM datasets and using the 15 minutes results for the rest of the datasets).
In general, we can say that there is a lot of false positives and that Dynotears tend to give a fully connected graph while constraint-based and hybrid based methods tend to give sparse graphs.

\begin{figure}[h]
	\centering
 	\begin{subfigure}{.4\textwidth}

    \caption{MVGCL}
    \end{subfigure}
    \hfill 
 	\begin{subfigure}{.4\textwidth}
	%
    \caption{Dynotears}
    \end{subfigure}

     \begin{subfigure}{.4\textwidth}
	%
    \caption{PCMCI$^+$}
    \end{subfigure}
    \hfill 
 	\begin{subfigure}{.4\textwidth}
	%
    \caption{PCGCE}
    \end{subfigure}

 	\begin{subfigure}{.4\textwidth}
	%
    \caption{VLiNGAM}
    \end{subfigure}
    \hfill 
 	\begin{subfigure}{.4\textwidth}
	%
    \caption{TiMINo}
    \end{subfigure}

     \begin{subfigure}{.4\textwidth}
	%
    \caption{NBCB-w}
    \end{subfigure}
    \hfill 
 	\begin{subfigure}{.4\textwidth}
	%
    \caption{NBCB-e}
    \end{subfigure}

         \begin{subfigure}{.4\textwidth}
	%
    \caption{CBNB-w}
    \end{subfigure}
    \hfill 
 	\begin{subfigure}{.4\textwidth}
	%
    \caption{CBNB-e}
    \end{subfigure}
	\caption{Inferred summary graph from MoM 1 dataset. Red edges correspond to false positive, black edges correspond to true positives and blue edges correspond to a true positive from one side and a false positive from another side.}
	\label{fig:controlled_infer_graph_1}
\end{figure}

\begin{figure}
	\centering
 	\begin{subfigure}{.4\textwidth}
	%
    \caption{MVGCL}
    \end{subfigure}
    \hfill 
 	\begin{subfigure}{.4\textwidth}
	%
    \caption{Dynotears}
    \end{subfigure}

     \begin{subfigure}{.4\textwidth}
	%
    \caption{PCMCI$^+$}
    \end{subfigure}
    \hfill 
 	\begin{subfigure}{.4\textwidth}
	%
    \caption{PCGCE}
    \end{subfigure}

 	\begin{subfigure}{.4\textwidth}
	%
    \caption{VLiNGAM}
    \end{subfigure}
    \hfill 
 	\begin{subfigure}{.4\textwidth}
	%
    \caption{TiMINo}
    \end{subfigure}

     \begin{subfigure}{.4\textwidth}
	%
    \caption{NBCB-w}
    \end{subfigure}
    \hfill 
 	\begin{subfigure}{.4\textwidth}
	%
    \caption{NBCB-e}
    \end{subfigure}

         \begin{subfigure}{.4\textwidth}
	%
    \caption{CBNB-w}
    \end{subfigure}
    \hfill 
 	\begin{subfigure}{.4\textwidth}
	%
    \caption{CBNB-e}
    \end{subfigure}
	\caption{Inferred summary graph from MoM 2 dataset. Red edges correspond to false positive, black edges correspond to true positives and blue edges correspond to a true positive from one side and a false positive from another side.}
	\label{fig:controlled_infer_graph_2}
\end{figure}

\begin{figure}
	\centering
 	\begin{subfigure}{.45\textwidth}
	%
    \caption{MVGCL}
    \end{subfigure}
    \hfill 
\begin{subfigure}{.45\textwidth}
	%
    \caption{Dynotears}
    \end{subfigure}

	\begin{subfigure}{.45\textwidth}
	%
    \caption{PCMCI$^+$}
    \end{subfigure}
    \hfill 
    	\begin{subfigure}{.45\textwidth}
	%
    \caption{PCGCE}
    \end{subfigure}

    	\begin{subfigure}{.45\textwidth}
	%
    \caption{VLiNGAM}
    \end{subfigure}
    \hfill 
    	\begin{subfigure}{.45\textwidth}
	%
    \caption{TiMINo}
    \end{subfigure}

        	\begin{subfigure}{.45\textwidth}
	%
    \caption{NBCB-w}
    \end{subfigure}
    \hfill 
    	\begin{subfigure}{.45\textwidth}
	%
    \caption{NBCB-e}
    \end{subfigure}

            	\begin{subfigure}{.45\textwidth}
	%
    \caption{CBNB-w}
    \end{subfigure}
    \hfill 
    	\begin{subfigure}{.45\textwidth}
	%
    \caption{CBNB-e}
    \end{subfigure}

	\caption{Inferred summary graph from Ingestion dataset. Red edges correspond to false positive, black edges correspond to true positives and blue edges correspond to a true positive from one side and a false positive from another side.}
	\label{fig:storm_infer_graph}
\end{figure}

\begin{figure}
	\centering
 	\begin{subfigure}{.45\textwidth}
	%
 
 \caption{CBNB-e}
    \end{subfigure}

	\caption{Inferred summary graph from Web 1 dataset. Red edges correspond to false positive, black edges correspond to true positives and blue edges correspond to a true positive from one side and a false positive from another side.}
	\label{fig:web_infer_graph1}
\end{figure}

\begin{figure}
	\centering
 	\begin{subfigure}{.45\textwidth}
	%
 
 \caption{CBNB-e}
    \end{subfigure}

	\caption{Inferred summary graph from Web 2 dataset. Red edges correspond to false positive, black edges correspond to true positives and blue edges correspond to a true positive from one side and a false positive from another side.}
	\label{fig:web_infer_graph2}
\end{figure}

\begin{figure}
	\centering
 	\begin{subfigure}{.45\textwidth}
	%
 
 \caption{CBNB-e}
    \end{subfigure}
	\caption{Inferred summary graph from Antivirus 1 dataset. Red edges correspond to false positive, black edges correspond to true positives and blue edges correspond to a true positive from one side and a false positive from another side.}
	\label{fig:antivirus_infer_graph1}
\end{figure}

\begin{figure}
	\centering
 	\begin{subfigure}{.45\textwidth}
	%
 
 \caption{CBNB-e}
    \end{subfigure}
	\caption{Inferred summary graph from Antivirus 2 dataset. Red edges correspond to false positive, black edges correspond to true positives and blue edges correspond to a true positive from one side and a false positive from another side.}
	\label{fig:antivirus_infer_graph2}
\end{figure}

\newpage

\subsection{Data examination}

Examination and visualization of time series is useful to observe trends, patterns, and dependencies in the data. By analyzing the data beforehand, we can identify potential behavior change, seasonality, sleeping time series, missing values or other time-dependent effects that may influence the outcomes we are interested in. Abnormal behavior, sleeping time series and misaligned data are a common occurrence in Monitoring data.

\subsubsection{MoM datasets}

Since MoM dataset was created in a controlled environment, the time series are aligned because sampling is uniformly collected at every second. There are also no missing values in this dataset, and no completely or partially sleeping time series.

\begin{figure}[h]
\centering
\includegraphics[width=1\textwidth]{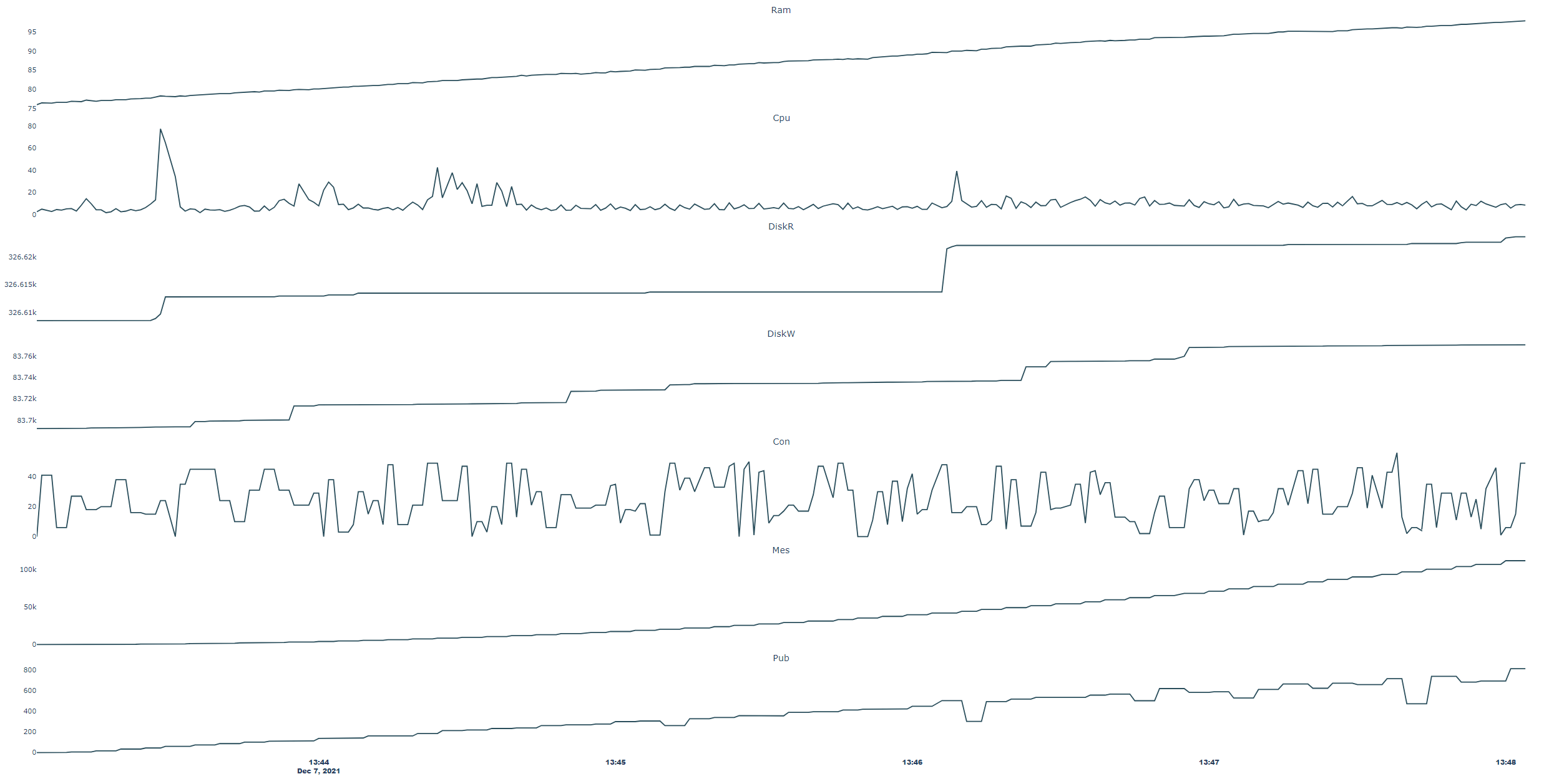}
\caption{Overview of MoM 1 dataset.}
\label{fig:mom1_raw_data}
\end{figure}

\begin{figure}[h]
\centering
\includegraphics[width=1\textwidth]{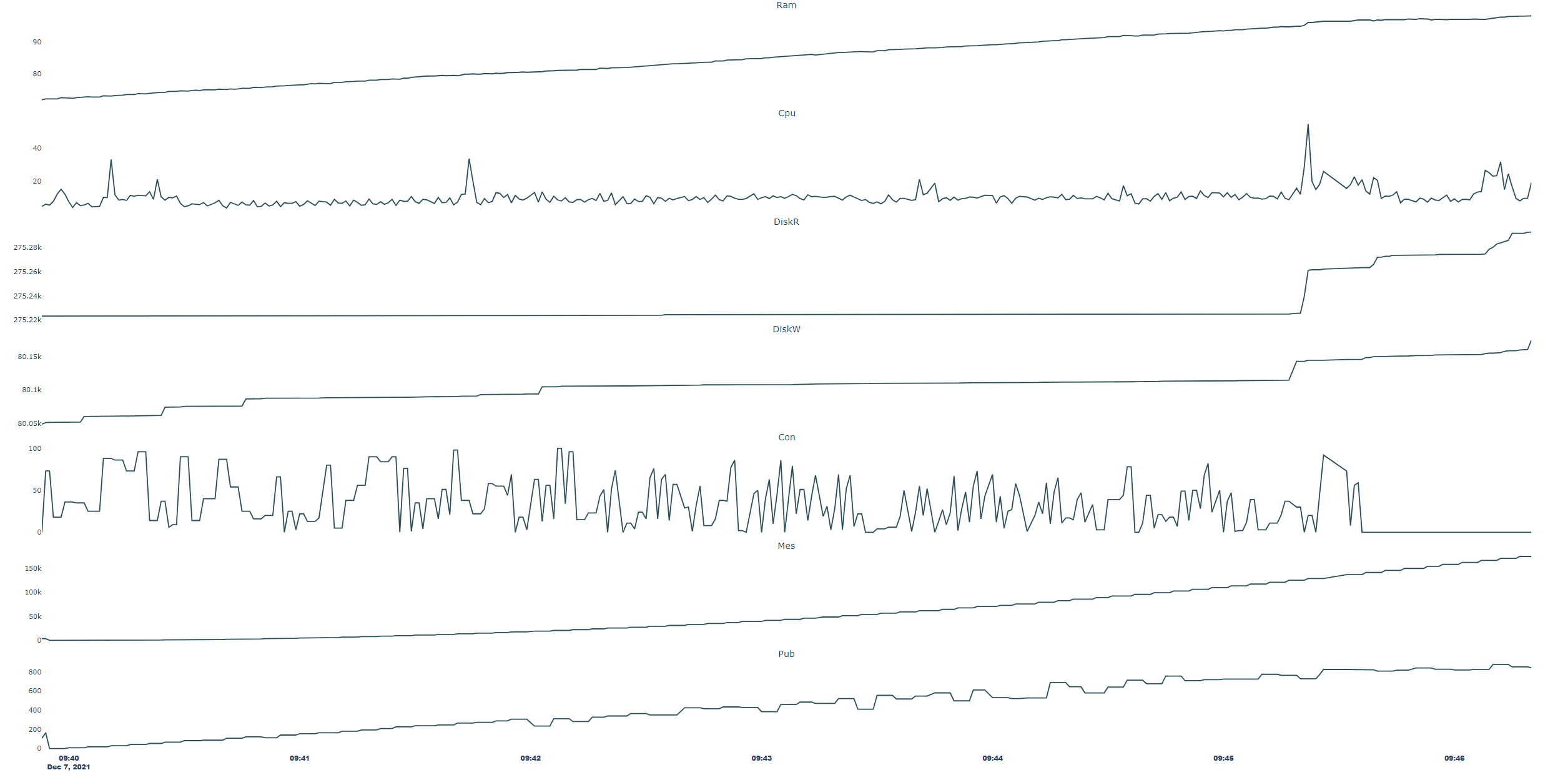}
\caption{Overview of MoM 2 dataset.}
\label{fig:mom2_raw_data}
\end{figure}

\subsubsection{Ingestion activity dataset}

\begin{figure}[h]
\centering
\includegraphics[width=1\textwidth]{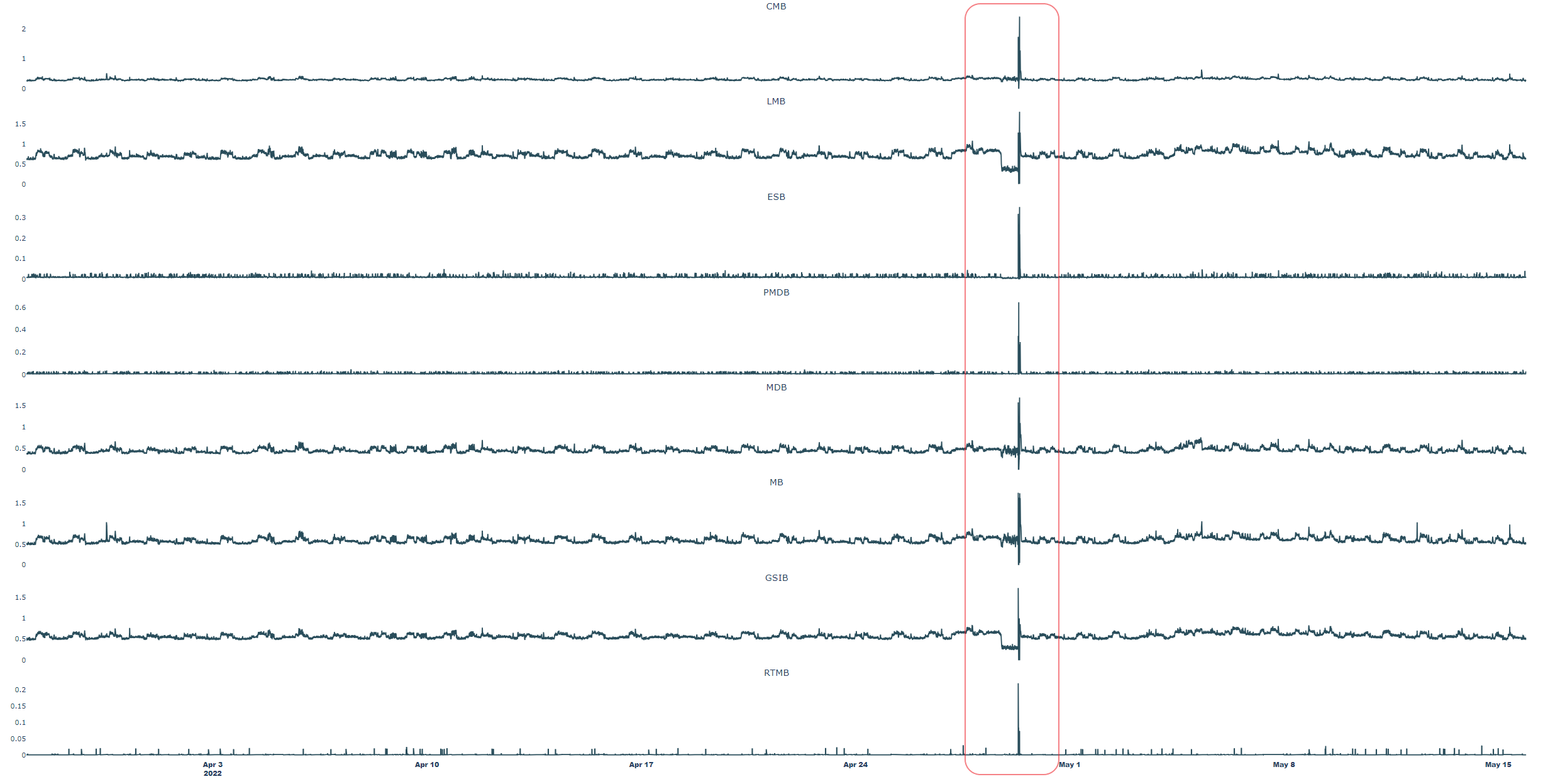}
\caption{Overview of Ingestion data, behavior change regions approximately highlighted inside the red box.}
\label{fig:ingestion_raw_data}
\end{figure}

As mentioned before, all the data are aligned for this dataset with sampling of $1$ minute. Moreover, it contained no missing values upon inspection. Figure~\ref{fig:ingestion_raw_data} contains a clear example of behavior change (highlighted in red). This is particularly interesting because the behavior change occurs in all time series approximately in the same region. There are no completely sleeping time series in this dataset, PMDB and RTMB are partially sleeping.

\subsubsection{Web activity dataset}

Upon examination of the $10$ time series, it was observed that the timestamps were not exactly aligned. It is noteworthy to mention that there were no sleeping time series observed in this dataset. However, NPP, NetIn and NetOut are partially sleeping. In terms of sampling, all time series had a sampling of $1$ minute. To align all the time series and make them of the same sampling, all the time series were resampled to $5$ minutes using either Strategy 1 or Strategy 2. Upon resampling, RamH and CpuG contained missing values, the maximum number of missing values was $1$ for both. The missing values were filled using simple linear interpolation of Pandas dataframes. It is important to mention that there were missing values in the raw data, but when sampled to a longer sampling the number of missing values were reduced. Afterwards, they were interpolated.

\begin{figure}[h]
\centering
\includegraphics[width=1\textwidth]{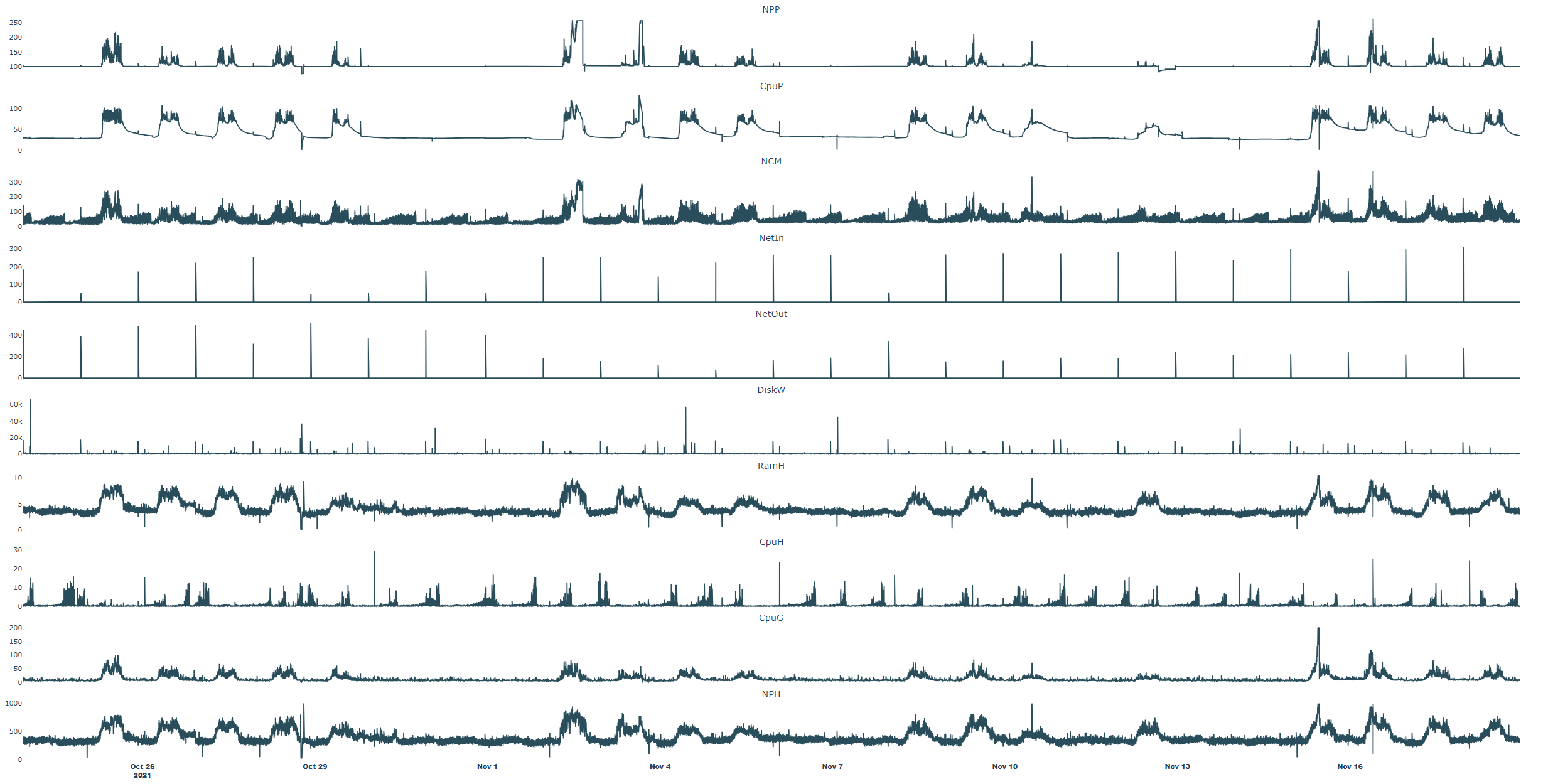}
\caption{Overview of raw Web data.}
\end{figure}

\begin{figure}[h]
\centering
\includegraphics[width=1\textwidth]{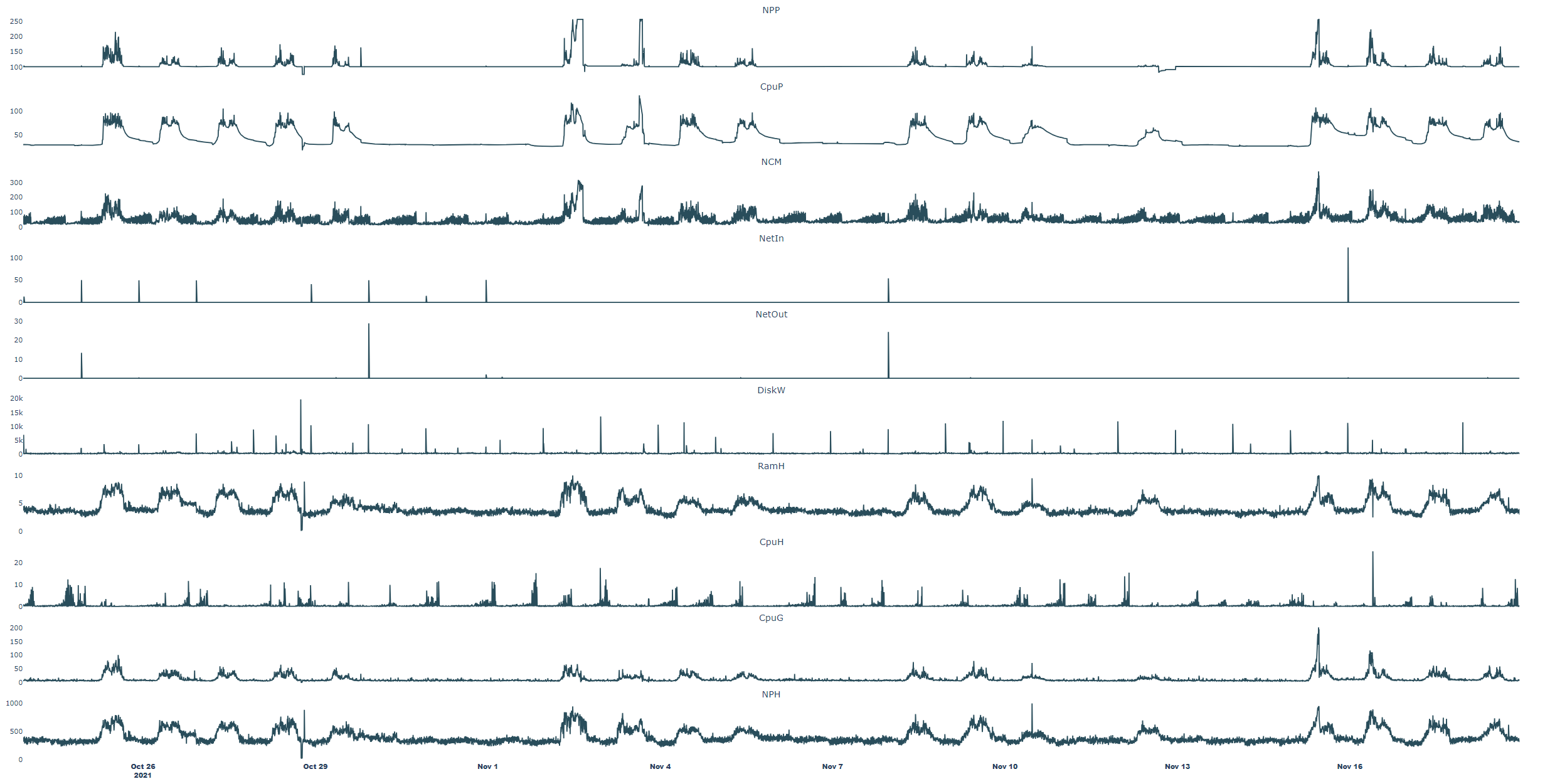}
\caption{Overview of Web data after pre-processing 1.}
\end{figure}

\begin{figure}[h]
\centering
\includegraphics[width=1\textwidth]{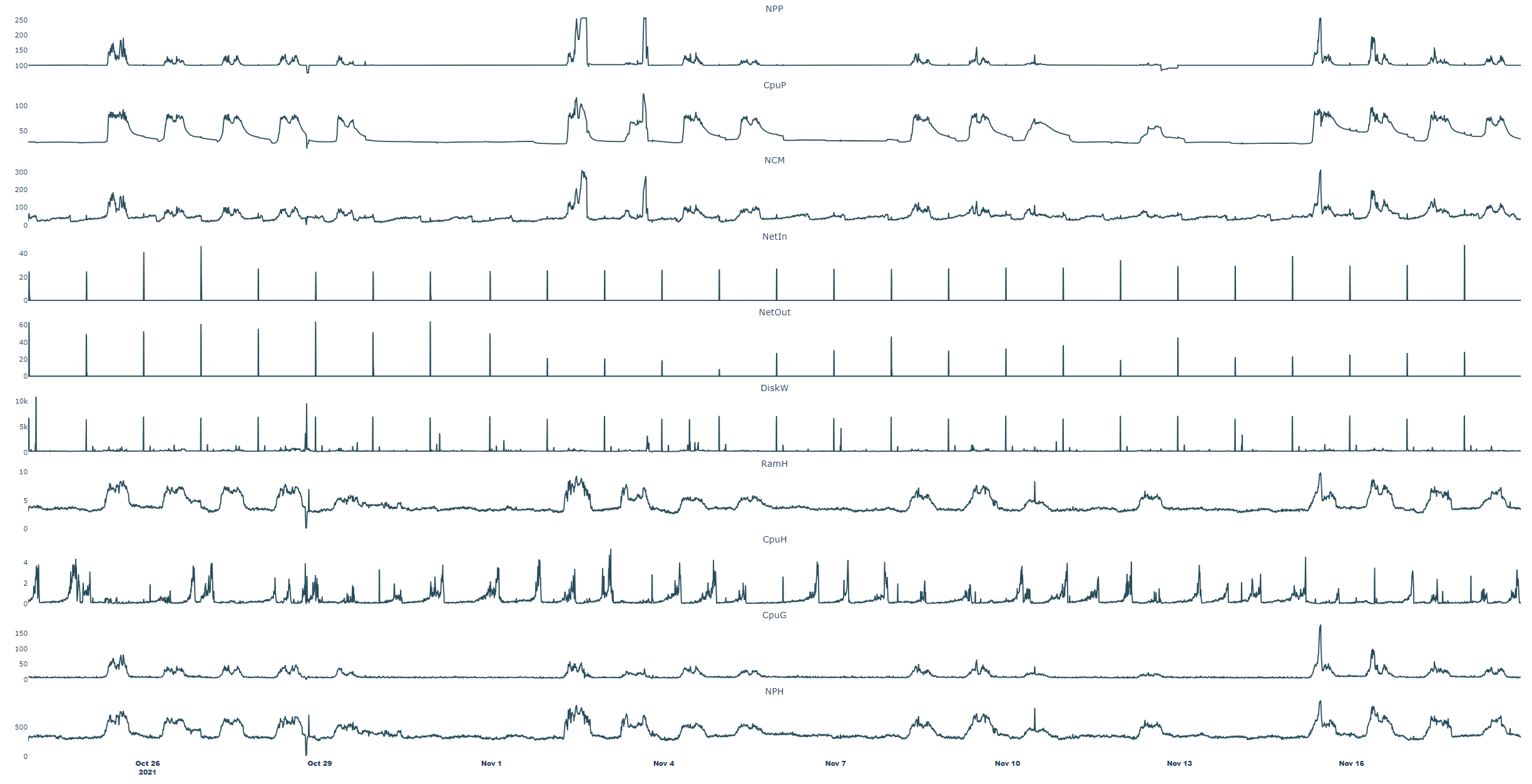}
\caption{Overview of Web data after pre-processing 2.}
\end{figure}

\subsubsection{Antivirus activity dataset}

This dataset contained $13$ time series in total, and the timestamps were not exactly aligned. Moreover, the raw data contained missing values. There were no completely sleeping time series observed in this dataset, but RP, CUP,RV,CUV and MUP were partially sleeping. In terms of sampling, ChIE, T and ChP had an original sampling of 5 minutes and the rest were $1$ minute. To align all the time series and make them of the same sampling, all the time series were resampled to $5$ minutes using either Strategy 1 or Strategy 2. Upon resampling, four metrics contained missing values. CUGV had $5$ missing values with at most $1$ consecutive missing value. CUV, ChP and MUP had $219$ missing values. However, there were at most $2$ consecutive missing values in these time series, so no large block of missing values was observed. The missing values were filled using simple linear interpolation of Pandas data frames. 

\begin{figure}[h]
\centering
\includegraphics[width=1\textwidth]{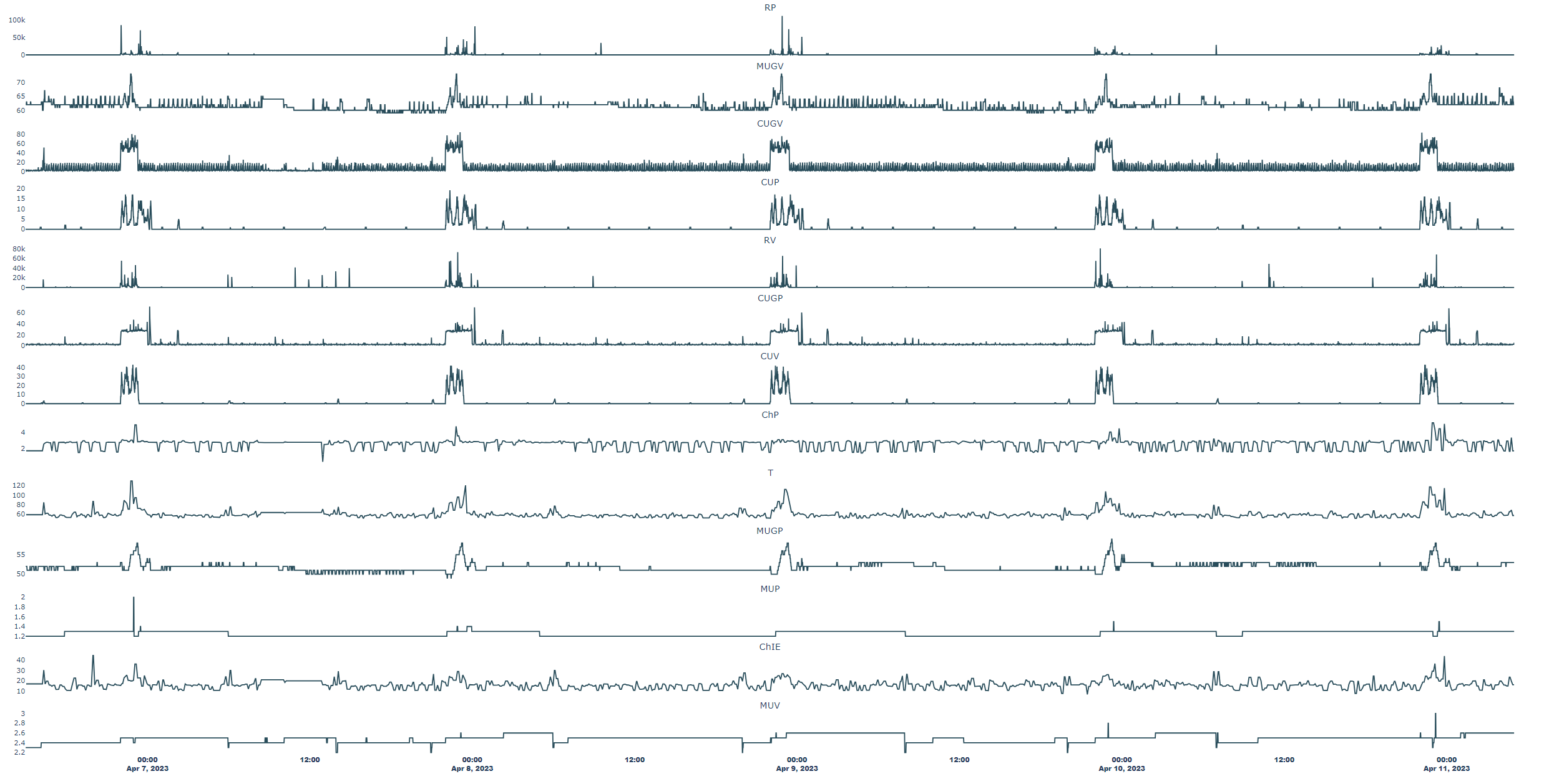}
\caption{Overview of raw Antivirus data.}
\end{figure}

\begin{figure}[h]
\centering
\includegraphics[width=1\textwidth]{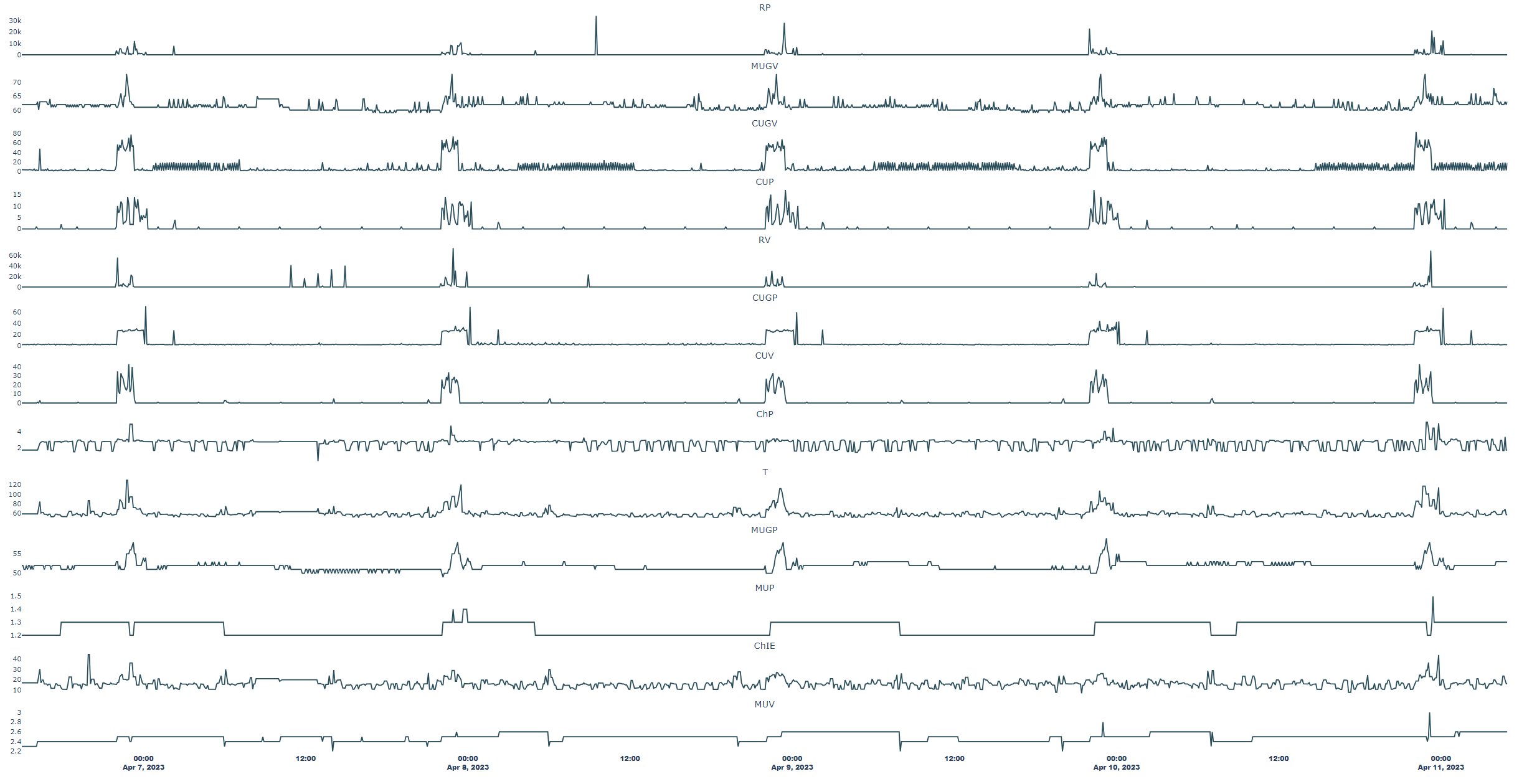}
\caption{Overview of Antivirus data after pre-processing 1.}
\end{figure}

\begin{figure}[h]
\centering
\includegraphics[width=1\textwidth]{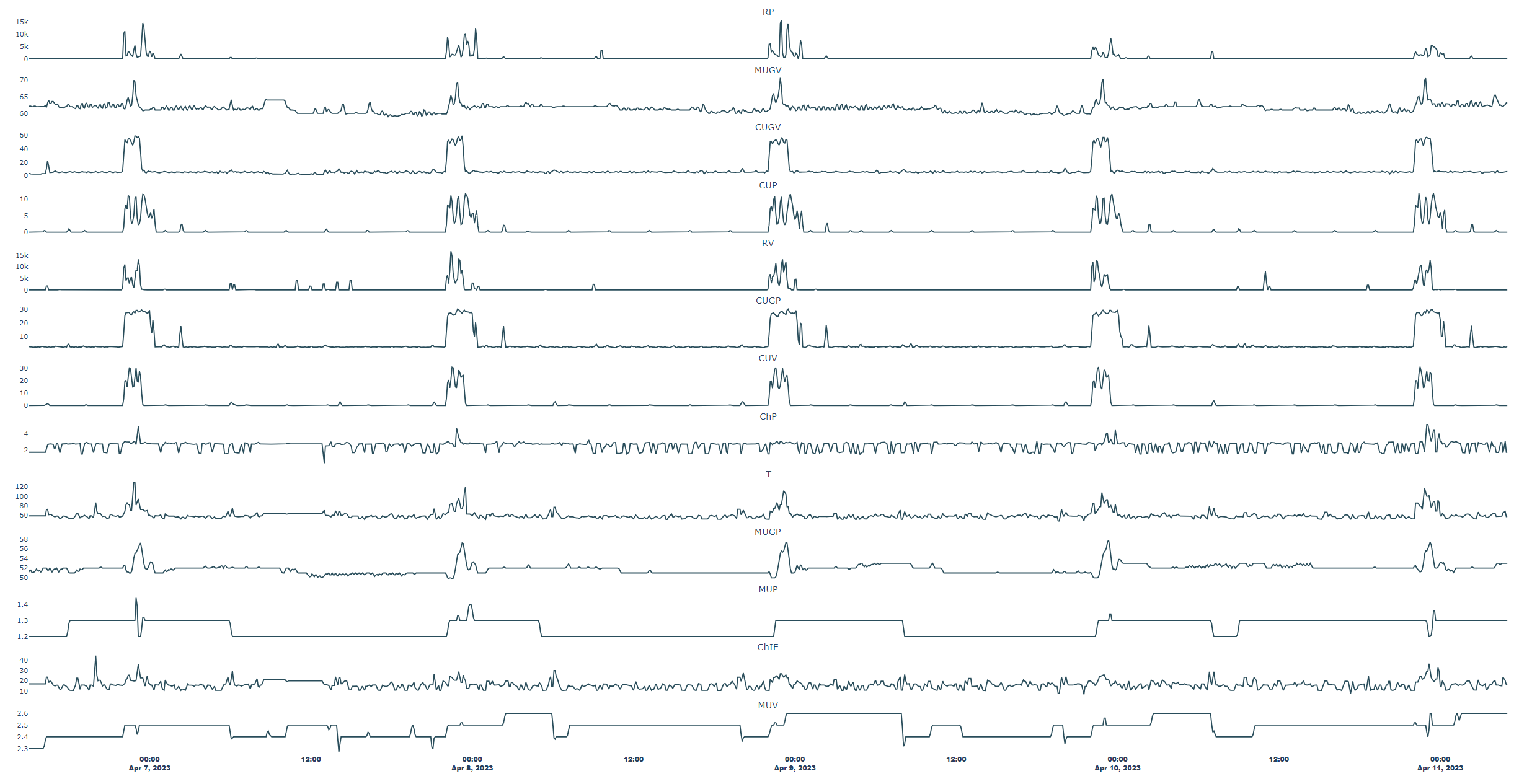}
\caption{Overview of Antivirus data after pre-processing 2.}
\end{figure}

\begin{table*}[ht]
	\caption{Summary of different datasets.} \label{tab:data_summary}
\centering
\begin{tabular}{r|c|c|c|c|c|c}
    \hline 
    & MoM  &  Ingestion  & Web & Antivirus\\ \hline
    Number sleeping time series after pre-processing& $0$ & $0$ & $0$ & $0$\\
    Number of partially sleeping time series after pre-processing & 0 & $2$ & $3$ & $5$ \\
    Sampling rate(s) before pre-processing & $1$ sec & $1$ min & $1$ min & $1$ \& $5$ mins \\
    Contained missing values before pre-processing & No & No & Yes & Yes\\
    Resampled after pre-processing & $1$ sec & $1$ min & $5$ mins & $5$ mins \\
    Number of time series with missing values after resampling & $0$ & $0$ & $2$ & $4$\\
\end{tabular}
\end{table*}

\end{document}